%% file: sn-article.tex
\documentclass[sn-mathphys-num,iicol]{sn-jnl}

\usepackage{graphicx}%
\usepackage{multirow}%
\usepackage{amsmath,amssymb,amsfonts}%
\usepackage{amsthm}%
\usepackage{mathrsfs}%
\usepackage[title]{appendix}%
\usepackage{xcolor}%
\usepackage{textcomp}%
\usepackage{manyfoot}%
\usepackage{booktabs}%
\usepackage{natbib}
\usepackage{algorithm}%
\usepackage{algorithmicx}%
\usepackage{algpseudocode}%
\usepackage{listings}%
\usepackage{pifont}
\usepackage{xcolor}
\definecolor{skyblue}{HTML}{0071BC}
\usepackage{color}
\usepackage{xspace}
\usepackage{overpic} 
\usepackage{multirow}
\usepackage{multicol}
\usepackage{booktabs}
\usepackage{bbding}
\usepackage{colortbl}
\usepackage{makecell}

\makeatletter
\DeclareRobustCommand\onedot{\futurelet\@let@token\@onedot}
\def\@onedot{\ifx\@let@token.\else.\null\fi\xspace}
\def\eg {\emph{e.g}\onedot} 
\def\ie{\emph{i.e}\onedot} 
 
\def\etc{\emph{etc}\onedot} 
 
\def\etal{\emph{et al}\onedot}
\makeatother

\theoremstyle{thmstyleone}%
\theoremstyle{thmstyletwo}%

\theoremstyle{thmstylethree}%

\begin{document}

\title[Article Title]{A Unified and Scalable Membership Inference Method for Visual Self-supervised Encoder via Part-aware Capability}


\author[1,2]{\fnm{Jie} \sur{Zhu}}\email{zhujie@stu.pku.edu.cn}
\author[3]{\fnm{Jirong} \sur{Zha}}\email{zhajirong23@mails.tsinghua.edu.cn}
\author[1,2]{\fnm{Ding} \sur{Li}}\email{ding\_li@pku.edu.cn}
\author*[1,2]{\fnm{Leye} \sur{Wang}}\email{leyewang@pku.edu.cn}

\affil[1]{\orgdiv{Key Lab of High Confidence Software Technologies (Peking University)}, \orgname{Ministry of Education}, \country{China}}

\affil[2]{\orgdiv{School of Computer Science}, \orgname{Peking University},  \state{Beijing}, \country{China}}

\affil[3]{\orgdiv{Tsinghua-Berkeley Shenzhen Institute}, \orgname{Tsinghua University}, \city{Shenzhen}, \country{China}}


\abstract{Self-supervised learning shows promise in harnessing extensive unlabeled data, but it also confronts significant privacy concerns, especially in vision. In this paper, we perform membership inference on visual self-supervised models in a more realistic setting: \textit{self-supervised training method and details are unknown for an adversary when attacking as he usually faces a black-box system in practice.} In this setting, considering that self-supervised model could be trained by completely different self-supervised paradigms, \eg, masked image modeling and contrastive learning, with complex training details, we propose a unified membership inference method called PartCrop. It is motivated by the shared part-aware capability among models and stronger part response on the training data. Specifically, PartCrop crops parts of objects in an image to query responses within the image in representation space. We conduct extensive attacks on self-supervised models with different training protocols and structures using three widely used image datasets. The results verify the effectiveness and generalization of PartCrop. Moreover, to defend against PartCrop, we evaluate two common approaches, \ie, early stop and differential privacy, and propose a tailored method called shrinking crop scale range. The defense experiments indicate that all of them are effective. Finally, besides prototype testing on toy visual encoders and small-scale image datasets, we quantitatively study the impacts of scaling from both data and model aspects in a realistic scenario and propose a scalable PartCrop-v2 by introducing two structural improvements to PartCrop. Our code is at \url{https://github.com/JiePKU/PartCrop}.}




\maketitle

\section{Introduction}\label{sec1}

Self-supervised learning~\cite{chen2020simple, chen2020big, jing2020self, gui2024survey} shows great potential in leveraging extensive unlabeled data.
While promising, self-supervised models are likely to involve individual privacy during training as they see massive data~\cite{chen2020simple} (\eg, personal or medical images) that are possibly collected from website without personal authorization~\cite{twitter_stop_url}.  
In addition, self-supervised models are usually used for services where users can feed their data and use its exposed API to extract data features (Internet giants like Google are already
offering ``machine learning as a service”~\cite{liu2021encodermi}.). It increases the privacy risk as adversaries~\footnote{In the rest of the article, without incurring the ambiguity, we use `adversary' to indicate person who manipulates attack and use `attacker' to indicate attack models an adversary utilizes.} may act as a normal person and feed some elaborate data that could lead to output containing sensitive privacy information~\cite{wei2022sparse, ren2024artificial}. Then adversaries can analyze these outputs and make privacy-related inference. 

In this research, we are interested in performing membership inference (MI) against self-supervised image encoders. Specifically, given an image, our goal is to infer whether it is used for training a self-supervised image encoder~\footnote{We primarily focus on membership inference in a black-box setting~\cite{Shaow_Learning, ResAdv} where only the output from the encoder is available as we deem that this setting is more likely to meet the realistic situation.}. Prior to our research, Liu~\etal have customized a membership inference method called EncoderMI~\cite{liu2021encodermi} for contrastive learning~\cite{caron2021emerging, he2020momentum}. However, self-supervised learning encompasses methods beyond contrastive learning, \eg, masked image modeling which has recently gained great prevalence~\cite{he2022masked}. Our experiment in Tab~\ref{tab:baseline result} shows that EncoderMI fails to attack models trained by masked image modeling. More importantly, EncoderMI assumes adversaries know the training details and employs the same data augmentations and hyperparameters as the target encoder. \textit{We contend that this setting is unjustified, as it might not accurately represent the true situation in real-world scenarios where self-supervised models are typically treated as black boxes~\cite{liu2021ml, Shaow_Learning, pre_conf_ML_Leaks, ResAdv}, particularly for Internet giants like Google.} In other words, we are only aware that the model within the black box is trained using one of the self-supervised methods, without knowledge of the specific method and its details~\cite{Shaow_Learning}~\footnote{In~\cite{Shaow_Learning}, it is said ``The details of the models and the training algorithms are
hidden from the data owners.''. We give a more detailed explanation in Sec~\ref{sec:background} Part 1.}. Hence, the practical application of EncoderMI may be limited.

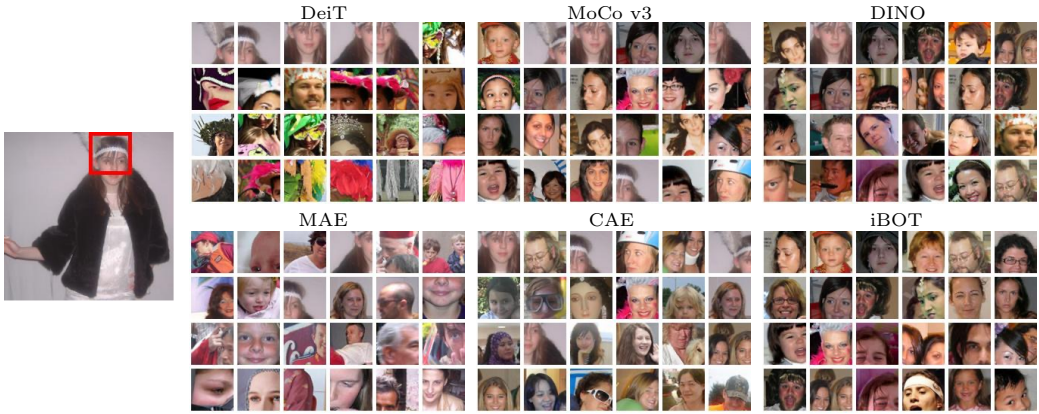
\begin{figure*}[t]
\vskip -0.45in
	\footnotesize
	\centering
        \begin{overpic}[width=.85\linewidth]{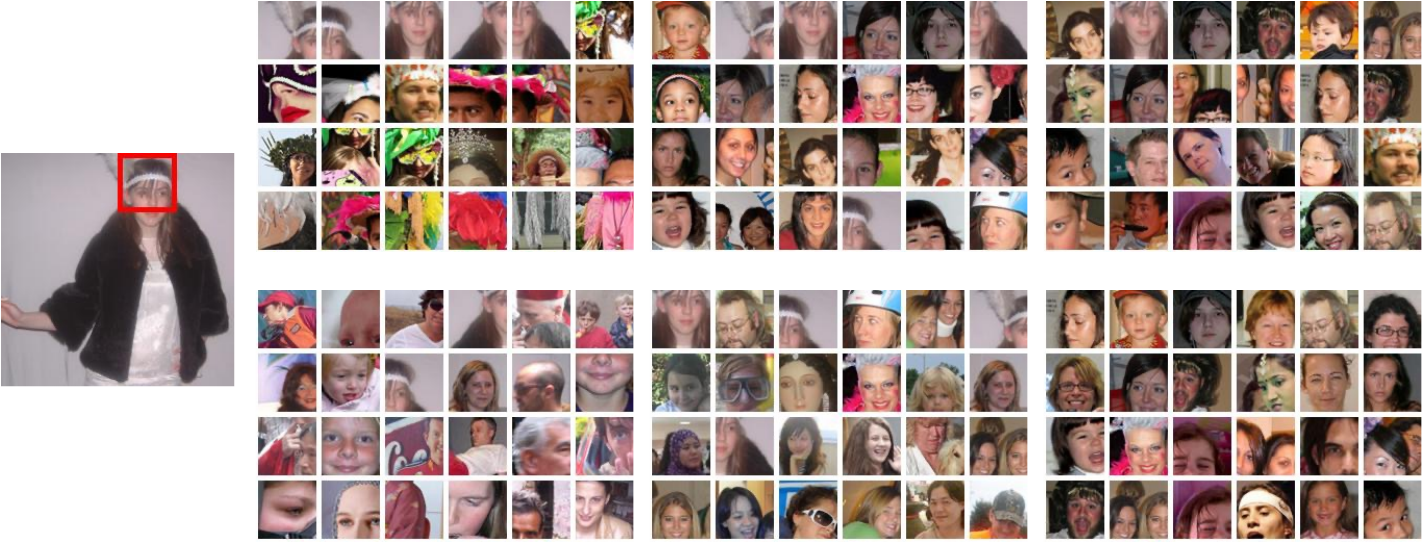}
    \put(28.7,38.4){DeiT}
    \put(54.5,38.4){MoCo v3}
    \put(83.8,38.4){DINO}
    \put(28.5,18.2){MAE}
    \put(56.6,18.2){CAE}
    \put(83.8,18.2){iBOT}
	\end{overpic}
	\caption{In a part retrieval experiment, the cropped part in red boundingbox (\ie, a human head) is used to query in a large image dataset. DeiT~\cite{touvron2021training} uses supervised learning. MAE~\cite{he2022masked} and CAE~\cite{chen2022context} are masked image modeling based methods.  DINO~\cite{caron2021emerging} and MoCo v3~\cite{chen2021empirical} are contrastive learning based methods. iBOT~\cite{zhou2021ibot} combines the two paradigms. This figure is borrowed from ~\cite{zhu2023understanding}. We refer readers of interest to \cite{zhu2023understanding}.} 
	\label{fig:part}
\end{figure*}



An intuitive solution to this circumstances is to systematically evaluate membership inference methods designed for different visual self-supervised methods and adopt the best-performing one. This would help adversaries overcome the limitation of lacking knowledge about the self-supervised training recipe. However, we have found that, to the best of our knowledge, there are no existing membership inference methods designed for visual self-supervised methods other than EncoderMI. This lack of alternatives can be attributed to the longstanding dominance of contrastive learning in the field of self-supervised learning, while a more promising paradigm, \ie, masked image modeling~\cite{he2022masked}, has only recently emerged.  
Moreover, even if we assume the existence of several inference methods, enumerating them all may be considerably intractable and inefficient due to the various pipelines and data processing involved, wasting substantial time.    

\textbf{To mitigate this dilemma, we aim to propose a unified membership inference method without prior knowledge of how self-supervised models are trained.} 
This setting involves two folds: First, self-supervised models could be trained by different paradigms, \eg, contrastive learning or masked image modeling~\footnote{
Considering the representative and prevalent of contrastive learning~\cite{chen2020simple, he2020momentum} and masked image modeling~\cite{bao2021beit, he2022masked}, they constitute our primary focus, while we also assess other three representative (or newly proposed) paradigms, illustrating the generalization of our method, see Sec~\ref{unified}};  Secondly, training details, \eg, data augmentations and hyperparameters, are also unknown for adversaries. 



\begin{figure}

\centering{\includegraphics[width=1\linewidth]{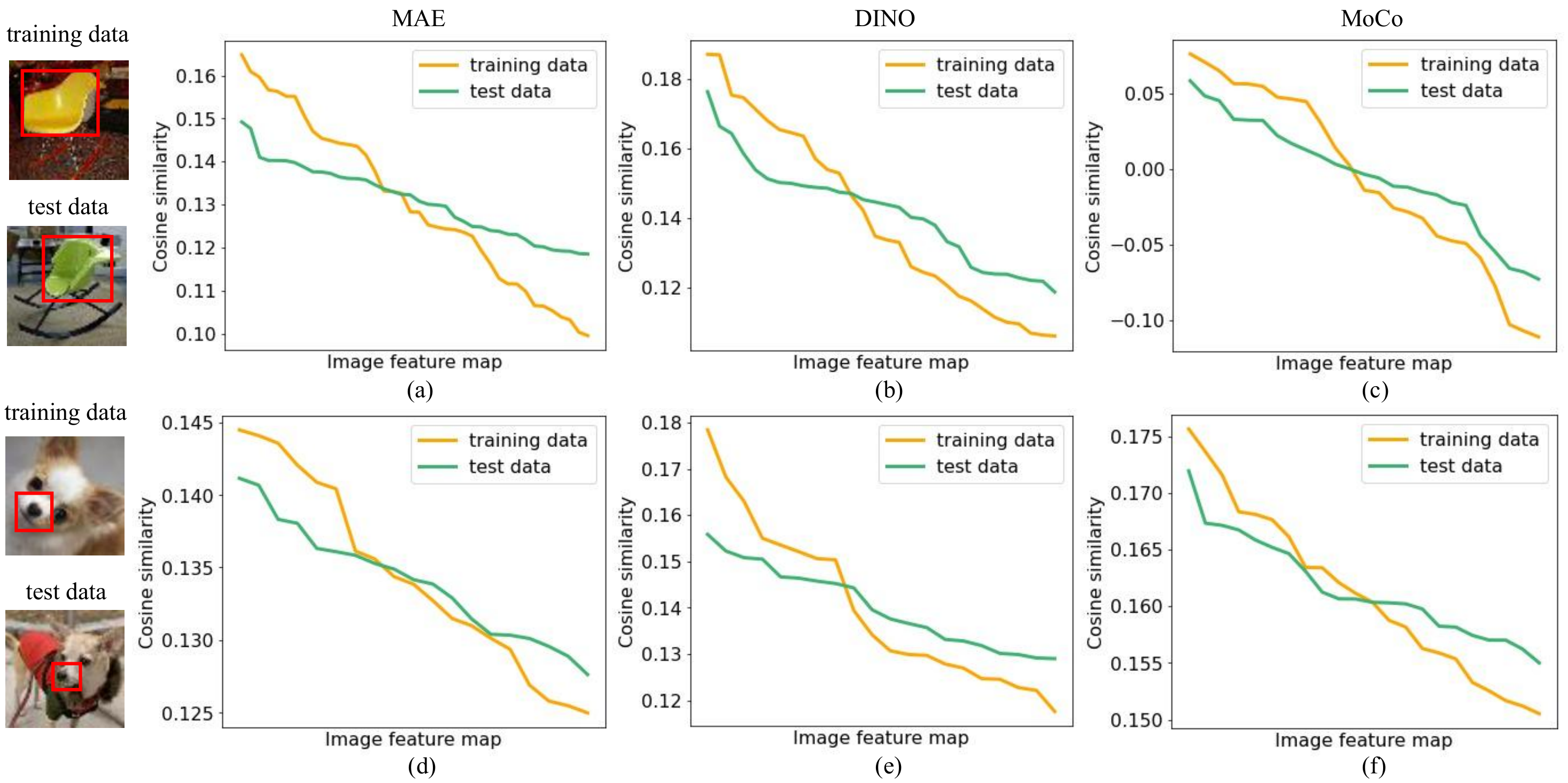}}
	\caption{Part response visualization on MAE (masked), DINO (contrastive), and MoCo (contrastive). Images are from Tinyimagenet~\cite{tinyimagenet_le2015tiny}. For each image containing a chair or dog, we manually crop it to obtain the \textit{part} (chair seat or dog muzzle) in the red box and resize it to suitable size. Next, we calculate the cosine similarity between each vector of the feature map of entire image and the part feature vector from the encoder, sorting the scores in descending order. (a), (b), (c) are the results of chair image and chair seat part on MAE, DINO, and MoCo, respectively. (d), (e), (f) are the results of dog image and dog muzzle part on MAE, DINO, and MoCo, respectively.}
	\label{fig:vis}
\end{figure}

Interestingly, we find that self-supervised models are capable of capturing characteristics of different \textbf{parts of an object~\footnote{The part is not simply a patch of an image. It means a part of an object, \eg, the head of a person, the mouth of a dog, and the eye of a cat.}}. As shown in Fig~\ref{fig:part}, self-supervised models like DINO and MAE basically retrieve various human heads while supervised method DeiT retrieves other parts, \eg, hair, ear, mouth, and clothes. This indicates that these self-supervised methods are superior at part perception (we refer to as ``part-aware capability"). This is also verified by concurrent work~\cite{zhu2023understanding}. Further, we are the first to find that training data exhibits a stronger part-aware capability compared to test data by showing their part similarity response in Fig~\ref{fig:vis}. Notably, the similarity curves for the training data are generally steeper than test data, indicating that part representations of training data are more discriminative than that of test data.

Motivated by this capability, we propose \textbf{PartCrop}. Given an image, it randomly crops numerous image patches to potentially yield as part-containing crops as possible~\footnote{Manually cropping the part of an object in an image is precise but inefficient, while automatic cropping requires sophisticated design. Hence, we adopt randomly cropping as a roundabout strategy.}. These patches, to some extent, can be considered as parts, and along with the global image, they are fed into a black-box self-supervised model to produce corresponding features. 
Then, part features serve as queries and interact with the image feature map to generate part-aware response. This response is regarded as \textit{membership feature} following~\cite{liu2021encodermi} and is collected as input of a simple inference attacker. Note that PartCrop is a \textit{unified} method as we do not specify the training methods and details of the given self-supervised model.  
Empirically, although the ``part" may not be as precise as manually cropped parts, PartCrop still significantly improves performance compared to random guessing (This may be the real result when adversaries are unaware of the training methods and details.). To defend against PartCrop, we evaluate two common methods, \ie, early stop and differential privacy, and propose a tailored defense method called shrinking crop scale range.

Finally, given that PartCrop has only been prototyped on small-scale image datasets and toy visual encoders, leaving a gap in real-world applicability, we scale both the dataset and model to more realistic scenarios. We systematically analyze how different scaling factors influence PartCrop's membership inference performance on self-supervised encoders, using the widely-adopted large-scale ImageNet1K dataset~\cite{imagenet15russakovsky}. Moreover, we find that PartCrop is relatively sensitive when scaling attacker and is likely to exhibit random guessing for MoCo. To alleviate this issue,  by visualizing the activation representation, we are inspired to introduce two improvements to attacker structure including normalization and activation function, resulting in a scalable attacker termed PartCrop-v2.



%

Our contribution can be summarized as follows:

$\bullet$ To the best of our knowledge, our work is the first to consider performing membership inference on self-supervised models without training recipe in hand. It is a more valuable setting in real life.

$\bullet$ We propose PartCrop, a unified method, by leveraging the part-aware capability in self-supervised models. In brief, PartCrop crops parts to measure the responses of image feature map and collect them as membership feature to perform inference. To defend against PartCrop, we consider two common approaches, \ie, early stop and differential privacy, and propose a tailored method called shrinking crop scale range. 

$\bullet$ Extensive experiments are conducted using three self-supervised models with different training protocols and structures on three computer vision datasets. The results verify the effectiveness of the attack and defense method.

$\bullet$ We quantitatively study the impacts of scaling from both data and model aspects under real-world conditions and further propose a scalable variant of PartCrop call PartCrop-v2 through introducing two structural improvements.

This article is an extension of a previous conference
edition~\cite{zhu2024unified}. Compared to the conference edition, this article
includes four main improvements: (1) studying how scaling data size influences the inference performance of PartCrop (2) analyzing how scaling model size affects the inference performance of PartCrop
(3) proposing a scalable PartCrop-v2  by introducing two structural improvements to PartCrop (4) involving more related work discussed in related work section.

\section{Background and Related Work}\label{sec:background}

\textbf{Unknowing the self-supervised training recipe for an adversary is a reasonable setting.} In reality, service systems are typically well-packaged, like black boxes~\cite{pre_conf_ML_Leaks, Shaow_Learning, ResAdv, liu2021ml}. Users input data and receive output after payment, without knowledge of the underlying model or its structure~\cite{Shaow_Learning}. This service mode is commonly used and widely adopted by internet giants such as Google and Amazon~\cite{liu2021ml}. This manner allows these companies to offer their services through simple APIs, thus making machine
learning technologies available to any customer~\cite{Shaow_Learning, ResAdv}. More importantly, it helps protect their core technologies from being stolen by adversaries~\cite{liu2021ml}. In the case of a self-supervised model system, the training recipe - which includes the self-supervised method used, data augmentation strategy, and hyperparameters - is critical for producing a successful self-supervised model for service. Therefore, it cannot be exposed to adversaries. Hence, when attacking self-supervised models, adversaries typically face a black box and are unaware of the used self-supervised method and details. From this view, this is a reasonable representation of the real-world setting.

\textbf{Why exploring scaling on membership inference for visual self-supervised learning?} Visual self-supervised learning demonstrates potential in processing extensive amounts of unlabeled images. This characteristic simplifies the scaling of data and facilitates the scaling of models even more, accelerating the presence of more general and intelligent vision system as Yann LeCun said~\cite{carrigan2021revolution}. However, the training images are often sourced from public sources, \eg, internet. With the increasing concerns about privacy and data security, understanding how scaling impacts the vulnerability of these models to membership inference is essential for safeguarding user privacy and helps in understanding the trade-offs between model performance and privacy risks. Finally, before deploying large visual self-supervised models in real-world applications, it is crucial to assess their vulnerability to privacy attacks. Understanding how scaling affects the susceptibility of these models to membership inference informs the deployment decisions and privacy-preserving measures~\cite{zhu2022safety, zhu2024safety}.

\textbf{Self-supervised Learning.} Self-supervised learning (SSL)~\cite{chen2020simple} is proposed to learn knowledge from substantial unlabelled data. Current self-supervised methods can be categorized into different families among which masked image modeling~\cite{bao2021beit, he2022masked, chen2022context, zhou2021ibot} and contrastive learning~\cite{chen2020simple, chen2020improved, chen2021empirical, caron2021emerging, grill2020bootstrap} are the most representative. Masked image modeling reconstructs masked part of an image via an encoder-decoder structure. Contrastive learning 
maximizes the agreement of the representations of random augmented views from the same image. Zhu~\etal~\cite{zhu2023understanding} show that masked image modeling and contrastive learning both have superior part-aware capability compared to supervised learning. In light of its amazing potential in leveraging massive unlabelled data, EncoderMI~\cite{liu2021encodermi} is the first membership inference method for contrastive-learning-based encoders. In this work, we aim to go further toward a more reasonable setting where an adversary does not know self-supervised training recipe, and propose a unified attack method dubbed as PartCrop.

\textbf{Membership Inference and Defense.} Prior research~\cite{carlini2019secret, nasr2019comprehensive} shows that a well-trained neural model is inclined to memorize the training data, resulting in different behavior on training data (members) versus test data (non-members)~\cite{hu2022membership}, \eg, giving a higher confidence (or stronger response) to training data. Hence, membership inference~\cite{Shaow_Learning, ResAdv} is proposed to infer whether a data record is used for training a neural model. When it comes to sensitive data, personal privacy is exposed to great risk, \eg, if membership inference learns that a target user's
CT images are used to train for a specific disease (\eg, to predict lung cancer~\cite{swensen2003lung}), then the adversary knows that the target user has the disease. Considering the privacy risks, it is valuable to study to defend against MI. Differential privacy~\cite{dp_abadi2016, dp_rahman2018, memguard_jia2019} adds noise to training data or models to disturb attackers. Nasr~\etal involve an attacker during training as adversarial regularization~\cite{ResAdv}. Zhu~\etal propose SafeCompress~\cite{zhu2022safety, zhu2024safety} to bi-optimize MI defense capability and task performance simultaneously when compressing models.

\textbf{Scaling Dataset \& Model.} Studies in
vision never stop exploring the impact of scaling model size and dataset. Russakovsky~\cite{imagenet15russakovsky} propose a large dataset including over one million images called ImageNet1K. This dataset marks a significant milestone in the advancement of visual intelligence. Then, a larger ImageNet21K~\cite{ridnik2021imagenet21k} is collected. He~\etal empirically demonstrate that masked autoencoders are
scalable learner~\cite{he2022masked} using ViT. More recently, Dehghani~\etal  scale the parameters of Vision Transformer~\cite{DosovitskiyB0WZ21} to 22 Billion~\cite{dehghani2023scaling}, showing the appealing potential of scaling.  These works collectively contribute to the understanding of how scaling datasets and model size can influence computer vision systems. Similarly, we scale membership inference to large self-supervised encoders and large-scale datasets instead of toy
models and datasets used in previous work~\cite{liu2021encodermi}.  We add more discussion about related work in App \ref{app:related work}.

\section{Threat Model}

For adversaries' ability, we consider a more realistic setting where adversaries have no knowledge of the target SSL model, \eg, training details, and only have black-box access to it. For threat model, following~\cite{liu2021ml}, we mainly consider two kinds of threat models for evaluation, \ie, \textbf{Partial} model and \textbf{Shadow} model. They are trained under the corresponding settings.


\textbf{Partial} setting means that an adversary may obtain partial training and test data of a target dataset, \eg, 50\%, due to data leakage, such as in the Facebook scandal~\cite{facebook_url}. In this situation, we assume that the adversary can directly use the partial data to train an attacker, and then attack the remaining data for evaluation.

\textbf{Shadow} setting indicates that an adversary has no knowledge of the training and test data of a target dataset. In this situation, the adversary has to resort to public datasets. Specifically, the adversary can first train a self-supervised model on a public dataset and then use this model and the public dataset to train an attacker. Afterward, the adversary uses this attacker to attack the target dataset. For example, the adversary can train an attacker on CIFAR100 and then perform an attack on a self-supervised model trained on Tinyimagenet. 

\textbf{Remark.} We regard attackers trained under partial setting as partial models (shadow models are the same). Considering that there could be numerous partial models (or shadow models) for different datasets, for simplicity, we use partial setting and shadow setting to indicate the type of threat model used in experiments. And we use partial setting if not specified.

\section{Method}
\begin{figure*}[t]
\vskip -0.45in
\centering{\includegraphics[width=1\linewidth]{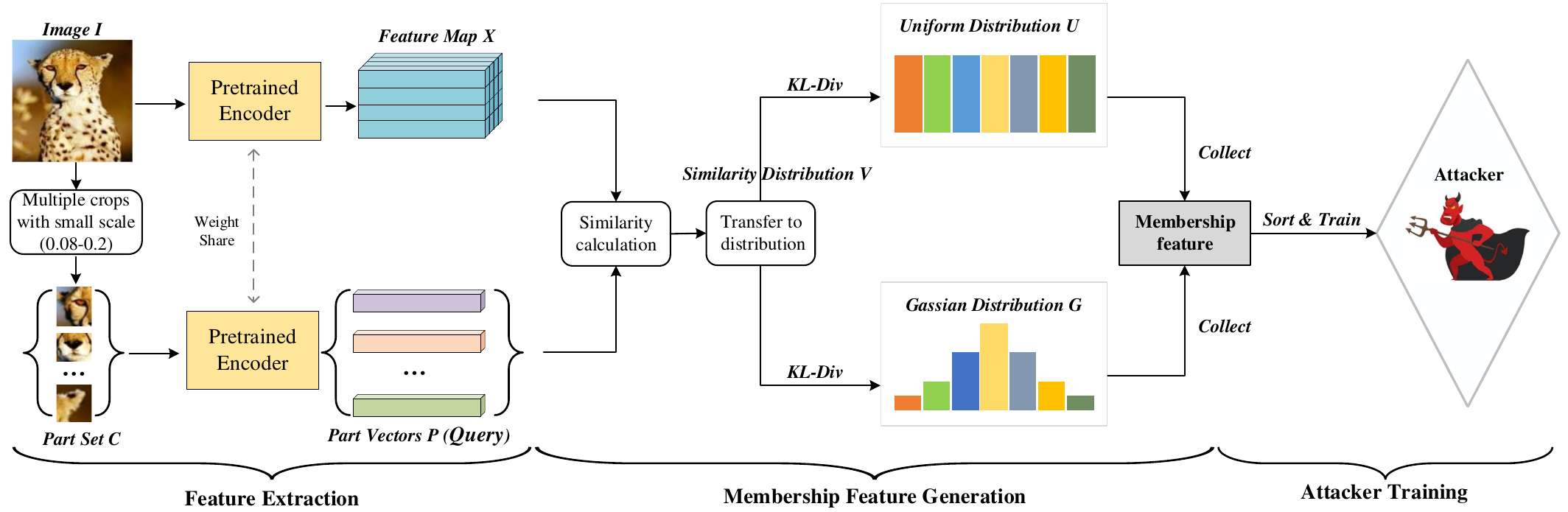}}
	\caption{An overview of PartCrop.}
	\label{overview}
 \vskip -0.1in
\end{figure*}

\textbf{Overview.} PartCrop contains three stages including feature extraction, membership feature generation, and attacker training as illustrated in Fig~\ref{overview}. Note that in this work, we primarily focus on image data and leave the application of PartCrop to other domains as future work.

\textbf{Stage 1: Feature Extraction.} Feature extraction is the first stage that contains two branches mapping image and part to the same representation space via a shared self-supervised encoder.

In the first branch, we extract the image feature. The whole image $I$ is fed into self-supervised model $\mathcal{F}$ that outputs a feature map $\chi \in \mathbb{R}^{H \times W \times D} $. The process is formulated as:
\begin{equation}
	\chi = \mathcal{F}(I) \,, \quad \chi \in \mathbb{R}^{H \times W \times D}
\end{equation}
$D$ is feature dimension. $H$ and $W$ are spatial size. We flatten the feature map along spatial dimension and reshape it as $\chi \in \mathbb{R}^{ N\times D}$ where $ N = H \times W $. 

In the second branch, we extract the part feature. We firstly randomly crop $m$ image patches that are likely to contain object parts from the whole image with crop scale $s$. We collect these crops in a set and denote it as $C = \{c_{0}, c_{1}, c_{2}, ......, c_{m-1}\}$ where $c_{i}$ is the $i$-th part crop. These crops then are fed into model $\mathcal{F}$ followed by avgpool and mapped to feature vectors in the same feature space as $\chi$. We denote these feature vectors as $P = \{p_{0}, p_{1}, p_{2}, ......, p_{m-1}\}$. We simply formulate it as:
\begin{equation}
P = avgpool(\mathcal{F}(C))
\end{equation}
For each element $p_{i}$ in $P$, $p_{i}$ is a feature vector with size $\mathbb{R}^{D}$. 

In our experiments, we set $m$ to $128$ and crop scale $s$ is set to $(0.08, 0.2)$ by default. We find that this setting is effective.

After finishing extracting features $\chi \in \mathbb{R}^{ N\times D}$ and $P = \{p_{0}, p_{1}, p_{2}, ......, p_{m-1}\}$ are sent to the second stage.

\textbf{Stage 2: Membership Feature Generation.}
Usually, in part-level perception, training data react more strongly than test data due to overfitting. Thus, in this stage, we regard each part crop representation vector $p_{i}$ in $P$ as a query to the feature map $\chi$ and return the response energy. And we consider the response energy as membership feature.

Given a query $p_{i}$, to obtain the response energy, we firstly calculate the similarity between query vector $p_{i}$ and feature map $\chi$ using the following formula:
\begin{equation} \label{eq:similarity}
\begin{split}
v_{i} = \chi \times p_{i} \quad \quad \quad \quad  \\ p_{i} \in \mathbb{R}^{D}\,, \; \chi \in \mathbb{R}^{ N\times D}\,, \; and \; v_{i} \in \mathbb{R}^{N}
\end{split}
\end{equation}
$\times$ indicates matrix multiplication. 

Now $v_{i}$ is the similarity vector between query $p_{i}$ and feature map $\chi$. An intuitive idea is to return the maximum value in $v_{i}$ as response energy. But this manner only focuses on local peak and ignores global similarity information. It could make membership feature not robust. Another method is to return the whole similarity vector. However, considering that we have $m$ queries, this could be somewhat over-size ($m\times N$) with potential redundance and wastes computational memory and resource for training an attacker.

Hence, we propose to leverage KL-divergence to integrate global similarity information.
Specifically, we transfer the global similarity vector to a similarity probability distribution following~\cite{vaswani2017attention} by using a softmax function:
\begin{equation}
\begin{split}
& v_{i}  = Softmax(v_{i}) \,, \\
& v_{ij} = \frac{e^{v_{ij}}}{\sum_{k=0}^{k=N-1} e^{v_{ik}}} \,, 
\end{split}
\end{equation}
where $v_{ij}$ is the $j$-th similarity value in $v_{i}$.

Then, to extract the similarity distribution characteristic, KL-divergence naturally stands out as it is designed to measure the difference between two distributions. Consequently, we use two common and independent distributions, \ie, uniform distribution $u_{i}$ and gaussian distribution $g_{i}$~\footnote{The gaussian distribution is randomly generated for each query. So we add a subscript $i$ to $u$ and $g$ for consistency. $u_{i}\sim U(0, D)$ and $g_{i}\sim N(0,1)$.}, as benchmarks, to evaluate the similarity distribution. The two distributions are easily implemented and morphologically different, introducing greater diversity in KL-divergence calculation. Empirically, we believe that the more the number of benchmark is, the richer the depicted similarity distribution characteristic is. In our experiment, we find that the two distributions are sufficient to achieve satisfying results.

In the end, our response energy $e_{i}$ for query $p_{i}$ is given by:
\begin{equation} \label{eq:response_energy}
\begin{split}
& e^{u}_{i} = KL\text{-}Div(v_{i} \; u_{i}) = \sum_{j=0}^{N-1} u_{ij}\log \frac{u_{ij}}{v_{ij}}   \,, \\
& e^{g}_{i} = KL\text{-}Div(v_{i}, \; g_{i}) = \sum_{j=0}^{N-1} g_{ij}\log \frac{g_{ij}}{v_{ij}} \,, \\
& e_{i} = [e^{u}_{i},\; e^{g}_{i} ] \,, 
\end{split}
\end{equation}
where $v_{ij}$, $u_{ij}$, and $g_{ij}$ are $j$-th  probability value in $v_{i}$, $u_{i}$, and $g_{i}$ respectively. 

As shown in Eq.~\ref{eq:response_energy}, our response energy $e_{i}$ contains two parts. The first is called uniform energy $e^{u}_{i}$ and the second is called gaussian energy $e^{g}_{i}$.

In our concrete implementation, we calculate the response energy for all queries in parallel. In this way, our method becomes more efficient and we can obtain the whole response energy $E = [E^{u},\; E^{g}],\; E \in \mathbb{R}^{ m\times 2}$ at once. Finally, we regard the acquired response energy $E$ as membership feature and use it to train a fully connected neural attacker in the next stage. 

\textbf{Stage 3: Attacker Training.} We use the extracted membership feature $E = [E^{u},\; E^{g}],\; E \in \mathbb{R}^{ m\times 2}$ to train an attacker $f$. It is an extremely simple fully connected network containing two MLP layers with ReLU as the activation function as shown in Tab~\ref{tab:attack structure}. 
We can also build a more complex attacker, \eg, add more layers and use self-attention module~\cite{vaswani2017attention}. However, our experiments indicate that a simple attacker performs well.  

To train the attacker, we firstly sort the uniform energy $E^{u} \in \mathbb{R}^{m}$ and gaussian energy $E^{g} \in \mathbb{R}^{m}$ by descending order respectively. Then, we concatenate $E^{u}$ and $E^{g}$ together and reshape it to a vector $E \in \mathbb{R}^{2m} $ as input. Finally, the attacker's output and corresponding label are fed into a loss function $\mathcal{L}$ for optimization. Note that we do not specify how the victim visual self-supervised encoder is trained and only use it to extract representation. Hence, PartCrop does not rely on the training prior of self-supervised encoder. 

\setlength{\tabcolsep}{0.8cm}\begin{table}[h]
	\centering
\footnotesize
\begin{tabular}{c | l}
		\toprule
		\textbf{Layers} & 
		\textbf{PartCrop}  \\
		\midrule
		Layer 1   & Linear(in\_dim, d) + ReLU() \\
  Layer 2   & Linear(d, d/2) + ReLU() \\
  Layer 3   & Linear(d/2, d/4) + ReLU() \\
  Layer 4   & Linear(d/4, 1) + Sigmoid() \\
		\bottomrule
	\end{tabular}
  \caption{The default attack structures of PartCrop. The $d$ determines the width of the attack structure. Here, the default width d is $512$. The in\_dim is the dimension of input feature and varies between different attack methods.}
  \label{tab:attack structure}
\end{table} 


\section{Experimental Setting}

\subsection{Self-supervised Model}
We use three famous and highly cited self-supervised models that represent combinations of different self-supervised paradigms and structures as shown in Tab~\ref{tab:models}.  They are MAE~\cite{he2022masked}, DINO~\cite{caron2021emerging}, and MoCo~\cite{he2020momentum}. We give a detailed introduction in App~\ref{app:Self-supervised Model}. 

\setlength{\tabcolsep}{0.2cm}\begin{table}[h]
	\centering
	\footnotesize
	\begin{tabular}{c c c}
		\toprule
		\textbf{\textit{Model}} & 
		\textbf{\textit{Self-supervised Method}} &
		\textbf{\textit{Structure}}  \\
		\midrule
		MAE & Masked Image Modeling & Vision Transformer \\
		DINO & Contrastive Learning & Vision Transformer \\
		MoCo & Contrastive Learning & CNN (ResNet) \\
		\bottomrule
	\end{tabular}
  \caption{Different self-supervised models.}
  \label{tab:models}
\end{table}

\subsection{Dataset and Metric}

We conduct experiments on three widely used vision datasets in membership inference field including CIFAR10, CIFAR100~\cite{cifar_10_krizhevsky2009learning}, and Tinyimagenet~\cite{tinyimagenet_le2015tiny}. For each dataset, the training/test data partition setting follows the reference papers.

\textbf{CIFAR10 and CIFAR100}~\cite{cifar_10_krizhevsky2009learning} are two benchmark datasets. Both of them have $50,000$ training images and $10,000$ test images. CIFAR10 has $10$ categories while CIFAR100 has $100$ categories. The dimension for CIFAR10 and CIFAR100 images is $32 \times 32 \times 3$. During self-supervised pretraining, we only use their training set.

\textbf{Tinyimagenet}~\cite{tinyimagenet_le2015tiny} is another image dataset that contains $200$ categories. Each category includes $500$ training images and $50$ test images. The dimension for Tinyimagenet images is $64 \times 64 \times 3$. Similar to CIFAR10 and CIFAR100, we also only use its training set for self-supervised training.

\textbf{Dataset Split.} For simplicity, we follow previous work~\cite{Pruning_IJCAI, shejwalkar2021membership, zhu2022safety}, which assumes a strong adversary that knows 50\% of the (target) model's training data and 50\% of the test data (non-training data) to estimate attack performance's upper bound. As shown in Tab~\ref{data_split} the known data are used for training the attacker while the remaining datasets are adopted for evaluation. We also evaluate other smaller ratios with a sweep from 10\% to 50\% in Varying Adversary's Knowledge part of App~\ref{app:Potential Tapping}.

\textbf{Metric.} During evaluation, we report attack accuracy, precision, recall, and F1-score (F1) as measurement following EncoderMI~\cite{liu2021encodermi}. And we pay more attention to attack accuracy in our experiments as the accuracy intuitively reflects the performance of attack methods on distinguishing member and non-member and treat them equally while attack precision, recall, and F1 primarily focus on member data. Simultaneously, we observe that if an attack method produces an accuracy close to 50\%, the precision and recall often vary irregularly. In this circumstances, the derived F1 is meaningless. Therefore, accuracy may serve as a more appropriate metric for evaluating attack performance and determining whether this method fails to attack (reduces to random guessing).  

\setlength{\tabcolsep}{0.1cm}\begin{table}[h]
	\centering	
	\footnotesize
	\begin{tabular}{l c  c | c  c}
		\toprule
		\multirow{2}{*}{\textbf{Datasets}}&
		\multicolumn{2}{c}{\textbf{Attack Training}}&\multicolumn{2}{c}{\textbf{Attack Evaluation}}\cr
		\cmidrule(lr){2-3} \cmidrule(lr){4-5}
		& $D^{known}_{train}$& $D^{known}_{test}$ &
		$D^{unknown}_{train}$ & $D^{unknown}_{test}$\\
		\midrule
		CIFAR10 & 25,000 & 5,000 & 25,000 & 5,000 \\
		CIFAR100 & 25,000 & 5,000 & 25,000 & 5,000 \\
		Tinyimagenet & 50,000 & 5,000 & 50,000 & 5,000 \\ 
		\bottomrule
	\end{tabular}
  \caption{Number of samples in dataset splits.}
  \label{data_split}
\end{table}

\subsection{Baseline}
To show the effectiveness of PartCrop, three baselines are carefully considered. The first is a repurposed supervised model based attack method~\cite{ResAdv, zhu2022safety}. It is chosen to show that this attack method is not suitable for self-supervised models. We denote this method as SupervisedMI. The second is repurposed from label-only inference attack~\cite{choquette2021label} called Variance-onlyMI. It leverages a variance prior. The third is a recently proposed attack method, \ie, EncoderMI~\cite{liu2021encodermi}, toward contrastive learning model conditioned on knowing the training recipe. It is a strong baseline and highly related to our work. We detail them in App~\ref{app:Baseline}.  

\setlength{\tabcolsep}{0.16cm}{\begin{table*}[h]
\vskip -0.45in
		\begin{center}
			\footnotesize
			\begin{tabular}{c| c | c c c c c c c c c c c c}
				\toprule
				\multirow{2}{*}{Method} & \multirow{2}{*}{Attacker} & \multicolumn{4}{c}{CIFAR100} & \multicolumn{4}{c}{CIFAR10}  & \multicolumn{4}{c}{Tinyimagenet} \cr
		\cmidrule(lr){3-6} \cmidrule(lr){7-10} \cmidrule(lr){11-14} 
                   & & Acc & Pre  &  Rec & F1 & Acc & Pre  &  Rec  & F1 & Acc & Pre  &  Rec & F1 \\
				\midrule
				\midrule
				\multirow{4}{*}{MAE} 
				 & SupervisedMI & 50.62 & 50.95	& 33.32 & 40.29 & 50.32 & 50.43 &	36.80 & 42.55 & 50.00&50.00 &20.04 & 28.61  \\
                    & Variance-onlyMI & 51.72 & 51.65	& 54.00	 & 52.80  & 51.52  & 51.60 &	49.16 & 50.35 & 50.10 & 50.10 & 50.82 & 50.46  \\
				 & EncoderMI & 51.70  & 52.22 & 49.88 & 51.02  & 53.20 & 53.18 &	61.06 & 56.85 & 50.18	& 52.19	 & 18.32 & 27.12 \\
				 & PartCrop &  \textbf{58.38}  & \textbf{57.73}	& \textbf{60.02} & \textbf{58.85} & \textbf{57.65}  & \textbf{57.60}	& \textbf{63.98} & \textbf{60.62} & \textbf{66.36}  &\textbf{62.68}	& \textbf{73.66}  & \textbf{67.73} \\
				\midrule 
				\multirow{4}{*}{DINO}
				& SupervisedMI & 49.60 &49.96&	12.04  & 19.40 & 50.50 & 50.52 & 47.82  & 49.13  & 50.34 &50.76&	0.66  & 1.30 \\
                    & Variance-onlyMI & 50.71	& 50.55	&  	\textbf{64.82}	& \textbf{56.80}   & 58.82 & 58.48 & 60.80	 &  59.62 & 55.81 & 54.25 & \textbf{74.12} & \textbf{62.65}  \\
				& EncoderMI & 55.52 &	57.20 &	44.82  &  50.26 & \textbf{66.40} & \textbf{65.55}& \textbf{63.60}  & \textbf{64.56} &  \textbf{63.87} & \textbf{66.73} &	54.96   & 60.27 \\
				& PartCrop & \textbf{60.62}  & \textbf{66.91} & 	41.08  & 50.91 & 59.13 &  60.66	& 53.76  & 57.00 & 56.13  & 62.72 &31.60  & 42.02 \\
				\midrule
				\multirow{4}{*}{MoCo}
				& SupervisedMI & 49.94	& 49.42	& 5.16  & 9.34 & 50.31 &51.05&10.62  & 17.58 & 50.21  &  54.07&1.46  & 2.84 \\
                    & Variance-onlyMI &  50.94  & 50.85 & 56.06  & 53.33  & 49.82  & 49.84 &	57.28 & 53.30 & 52.66 & 52.80 & 50.22 & 51.48  \\
				& EncoderMI & 57.29 &	58.14 &	59.12  & 58.62 & 60.81 &	58.75 &	\textbf{72.24}  & 64.80 & 62.33 &	70.29&	41.26  &  51.99 \\
				& PartCrop & \textbf{77.20} & \textbf{82.93} &	\textbf{67.84}   & \textbf{74.63} & \textbf{78.84}  & \textbf{86.46} & 66.94  & \textbf{75.46} &\textbf{73.77} & \textbf{76.02} & 	\textbf{69.20}  &  \textbf{72.45} \\
				\bottomrule
			\end{tabular}
		\end{center}
        \vskip 0.05in
       \caption{Comparisions with SupervisedMI, Variance-onlyMI, and EncoderMI in partial setting. Acc: Accuracy, Pre: Precision, Rec: Recall} 
       \label{tab:baseline result}
\end{table*}}

\subsection{Implementation Details}\label{sec:Implementation Details}

We use official public code for self-supervised training~\footnote{MAE: \url{https://github.com/facebookresearch/mae.}}~\footnote{DINO: \url{https://github.com/facebookresearch/dino.}}~\footnote{MoCo: \url{https://github.com/facebookresearch/moco}}. Concretely, they are pretrained for 1600 epochs following their official setting including data augmentation strategies, learning rate, optimizers, \etc. Due to limited computational resources, we use Vision Transformer (small)~\cite{DosovitskiyB0WZ21} and set patch size to 8 for MAE and DINO. We use ResNet18~\cite{resnet_he2016deep} as backbone of MoCo.
We set batch size to 1024 for all models during pretraining. In PartCrop, for each image, we produce 128 patches that potentially contain parts of objects. Each patch is cropped with a scale randomly selected from 0.08-0.2. Afterward, we resize the part to 16 $\times$~16. When training attacker, following previous work~\cite{Shaow_Learning, ResAdv,zhu2022safety}, all the network weights are initialized with normal distribution, and all biases are set to 0 by default. The batch size is 100. We use the Adam optimizer with the learning rate of 0.001 and weight decay set to 0.0005. We follow previous work~\cite{liu2021ml, ResAdv} and train the attack models 100 epochs. During training, it is assured that every training batch contains the same number of member and non-member data samples, which aims to prevent attack model from being biased toward either side.

\section{Experiment}\label{sec:experiment} \vskip -0.05in

\subsection{Results in the Partial Setting}\label{partial}
To evaluate the effectiveness, we compare PartCrop, with SupervisedMI, Variance-onlyMI, and EncoderMI. We use the \textit{partial setting} where an adversary knows part of training and test data and report the results in Tab~\ref{tab:baseline result}. 

SupervisedMI achieves approximately 50\% attack accuracy across all scenarios and the precision and recall vary irregularly, suggesting its inability to attack self-supervised models trained on any of the datasets.  
It is reasonable as SupervisedMI is designed to attack supervised model. Generally, SupervisedMI should take supervised models' category probability distribution (in label space) as input while self-supervised models'output is encoded image feature (in representation space). This space gap leads to the failure of SupervisedMI.

Similarly, Variance-onlyMI fails across most scenarios while only performing well on DINO trained by CIFAR10 and Tinyimagenet. Also, its inference performance is inferior to EncoderMI and PartCrop. These results indicate that for Variance-onlyMI, utilizing variance as a characteristic for membership inference doesn't work for all SSL models.

EncoderMI performs well on MoCo and DINO, and even outperforms PartCrop in some cases. For example, EncoderMI outperforms PartCrop in Tinyimagenet by around 8\% on accuracy, 4\% on precision. However, when performing on MAE, a masked image modeling based model, the attack accuracy of EncoderMI drops significantly (CIFAR100 and CIFAR10) and even reduces to random guess (Tinyimagenet). As for precision and recall, they decrease accordingly. The derived F1 is also lower than that of PartCrop. This contrast of results implies that EncoderMI heavily relies on the prior of contrastive learning and the training recipe of victim model to obtain satisfying performance. In the case of masked image modeling, which aims to reconstruct the masked portions, it is differs from contrastive learning. Consequently, EncoderMI naturally fails to obtain satisfying results on MAE. Finally, we consider adapting EncoderMI to attack MAE, \eg, replacing augmented views generated with patches. We conduct this experiment on CIFAR100. The adaptation achieves 51.18 for accuracy, inferior to the original performance (51.70 Acc), further demonstrating the essential discrepancy between masked image modeling and contrastive learning.

In contrast, without knowing how self-supervised model is trained, PartCrop generally achieves satisfying attack performance on four metrics for all the models, which greatly demonstrates the effectiveness of our PartCrop. 


\subsection{Ablation Study}
We ablate three key elements in PartCrop including membership feature, crop number, and crop scale on CIFAR100 dataset, revealing how they influence the membership inference performance of our PartCrop, respectively. We primarily focus on the accuracy and also use F1 to represent the trade-off between precision and recall.

\setlength{\tabcolsep}{0.42cm}{\begin{table}[h]
\vskip -0.45in
			\footnotesize
			\begin{tabular}{c c c c c}
				\toprule
				Model & $E^{u}$ & $E^{g}$ & Accuracy & F1 \\ 
				\midrule
				\midrule
				\multirow{3}{*}{MAE} 
				& \Checkmark  & & 57.43 &  54.98 \\
				&  & \Checkmark & 56.67 &	58.18 \\
				&  \Checkmark &  \Checkmark &  \textbf{58.38} &  \textbf{58.85} \\
				\midrule 
				\multirow{3}{*}{DINO}
				& \Checkmark & & 54.16  &   48.39 \\
				&  & \Checkmark& 51.56 &  48.94 \\
				&\Checkmark &  \Checkmark &\textbf{ 60.62}  & \textbf{50.91} \\
				\midrule
				\multirow{3}{*}{MoCo}
				& \Checkmark & & 70.27 &  68.97 \\
				&  & \Checkmark & 69.49 & 69.15 \\
				& \Checkmark & \Checkmark &\textbf{ 77.20} & \textbf{74.63} \\
				\bottomrule
			\end{tabular}
      \caption{Ablation study on membership feature. $E^{u}$ is uniform energy and $E^{g}$ is gaussian energy. \Checkmark $\;$ means using the energy during training and evaluating attacker.} 
         \label{tab:membership feature}
		\vspace{-8pt}  
\end{table}}

\subsubsection{Merbership Feature} In PartCrop, membership feature contains uniform energy $E^{u}$ and gaussian energy $E^{g}$. We investigate the impact of these factors separately and jointly on the attack performance. As shown in Tab~\ref{tab:membership feature},  $E^{u}$ and $E^{g}$ are both useful for membership inference. Moreover, we find that $E^{u}$ leads to higher attack accuracy than $E^{g}$ consistently among three models. For example, in MAE, $E^{u}$ produces 57.43\% for accuracy while $E^{g}$ is inferior and achieves 56.67\% accuracy. However, this situation is reversed when it comes to F1. When leveraging both $E^{u}$ and $E^{g}$, it further improves the attack performance on accuracy and F1.  This indicates that $E^{u}$ and $E^{g}$ are complementary to some extent. 

\subsubsection{Crop Number}
\begin{figure}
\vskip -0.45in
\centering{\includegraphics[width=0.9\linewidth]{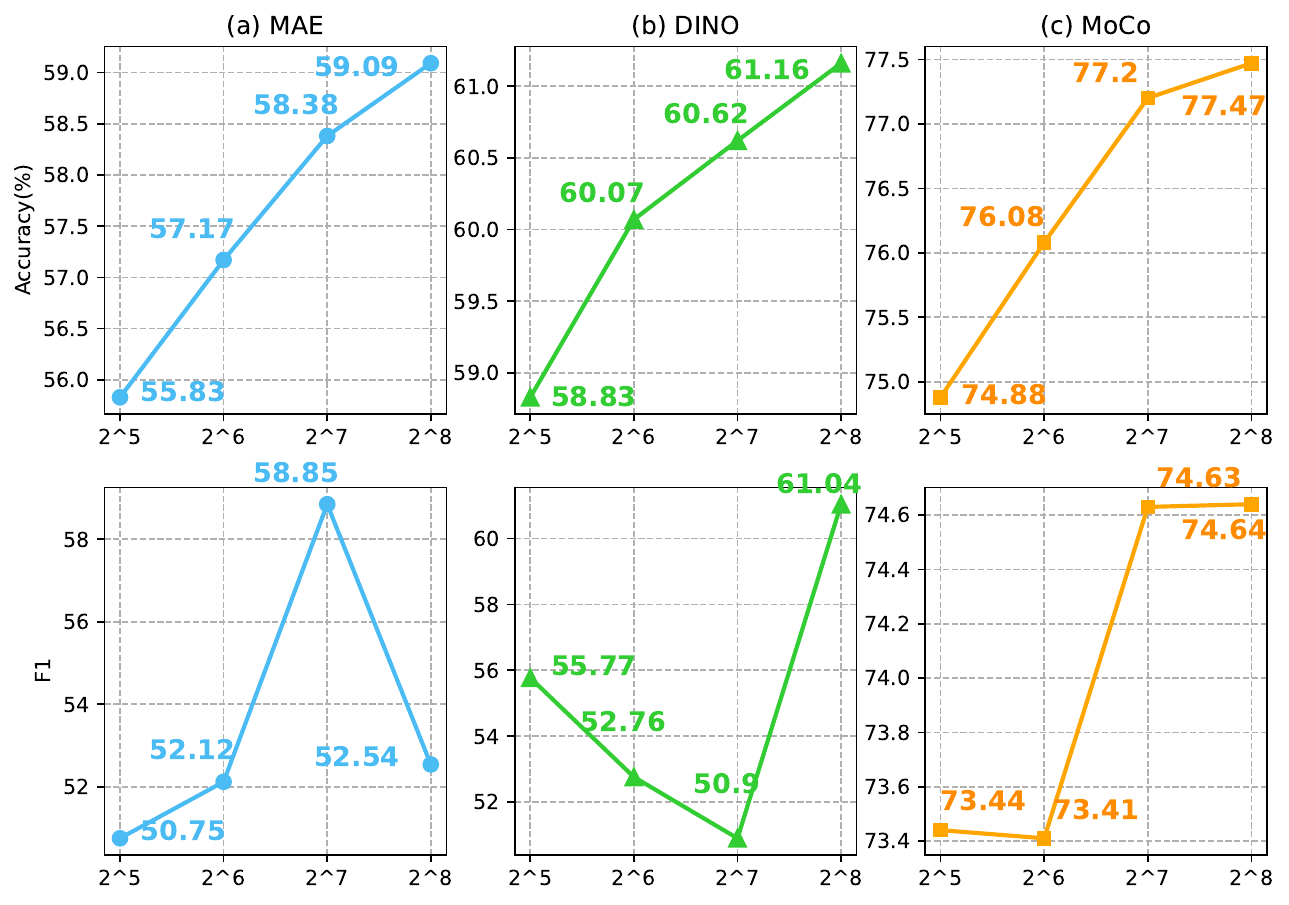}}
	\caption{Ablation study on crop number. We consider four different crop number \ie, 32, 64, 128, and 256.}
	\label{fig:crop number}
\end{figure}

The crop number in PartCrop is critical. It determines how many patches we use. In this experiment, we consider four different crop numbers \ie, $32$, $64$, $128$, and $256$. Fig~\ref{fig:crop number} shows the relationship between query number and attack performance. Generally, more queries lead to better accuracy but more computation. Moreover, after a certain point, increasing crop numbers brings only marginal improvement. In terms of F1, it exhibits opposite behavior, especially for MAE and DINO, indicating that crop number has different impact on F1, possibly because of the mutual competition between precision and recall. Considering that increasing crop number aggravates the computational burden, we select a trade-off (by using $128$ as our default setting) to ensure high attack performance while minimizing computation cost.

\subsubsection{Crop Scale}\label{sec:crop_scale}
\begin{figure}[t]
\centering{\includegraphics[width=0.95\linewidth]{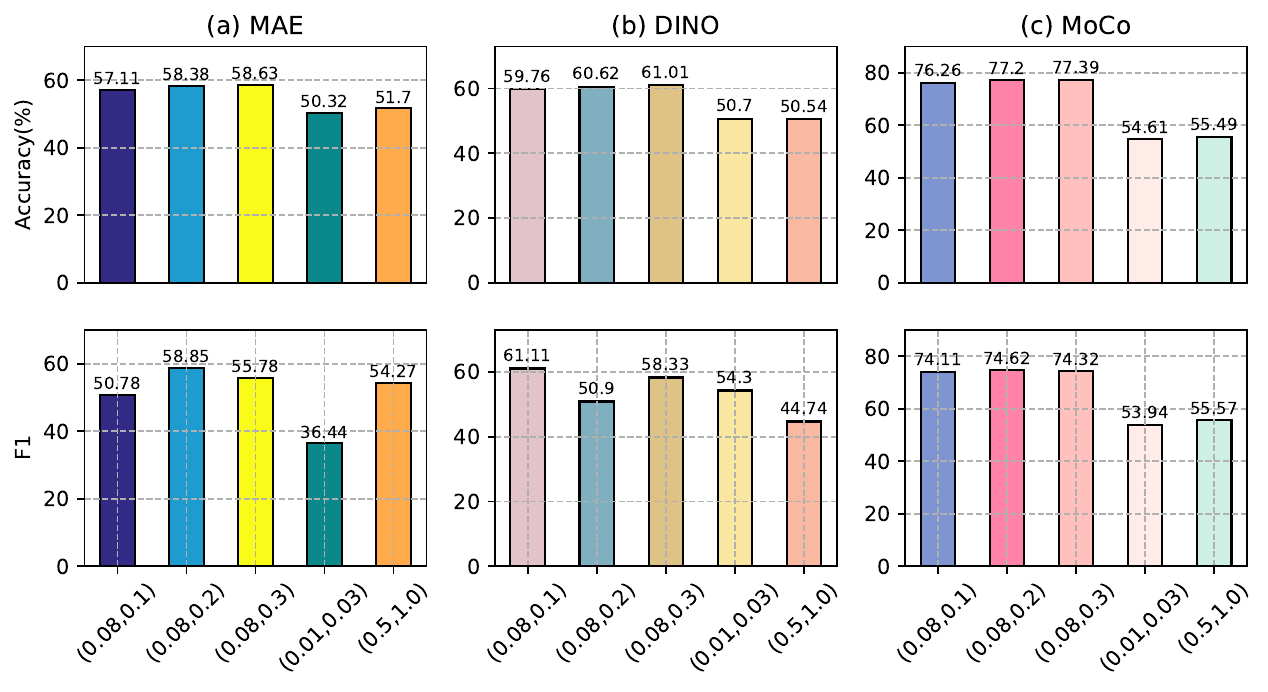}}
	\caption{Ablation study on crop scale. We consider five different crop scales \ie, $(0.08,\; 0.1)$, $(0.08,\; 0.2)$, $(0.08,\; 0.3)$, $(0.01,\; 0.03)$, and $(0.5,\; 1.0)$.}
	\label{fig:crop scale}
\end{figure}

In PartCrop, crop scale determines how much ratio the area of the cropped patch is in that of raw image. We consider three crop scales setting including $(0.08,\; 0.1)$, $(0.08,\; 0.2)$, and $(0.08,\; 0.3)$. In Fig~\ref{fig:crop scale} shows that from $(0.08,\; 0.1)$ to $(0.08,\; 0.2)$, PartCrop produces about 1\% improvement for attack accuracy among three models.  As for F1, the influence is inconsistent: it significantly enhances the F1 of MAE; In contrast, it significantly reduces the F1 of DINO. Compared to DINO and MAE, MoCo undergoes mild F1 changes. We conjecture that this could be caused by the structure discrepancy: MAE and DINO are Vision Transformer while MoCo is CNN. 
While in scale $(0.08,\; 0.3)$, compared to $(0.08,\; 0.2)$, the attack accuracy is marginal, especially for MoCo. Hence, we use $(0.08,\; 0.2)$ as our default crop scale.

We also add two particular crop scales, \ie, $(0.01,\; 0.03)$ and $(0.5,\; 1.0)$,  to study the impact of extremely small and large crop scale. As shown in Fig~\ref{fig:crop scale}, $(0.01,\; 0.03)$ fails in membership inference for MAE and DINO and drops significantly for MoCo in terms of accuracy. It is reasonable as such scale makes the patches contain little useful part information. For another scale  $(0.5,\; 1.0)$, it causes the cropped image to cover most of the area of raw image, possibly including the whole object (in this situation, we can not call it part.). This violates our motivation of leveraging part capability in self-supervised models as the input is not a part of an object but more likely the whole. Consequently, the attack accuracy drops significantly. For F1, MAE obtains $36.44$ in $(0.01,\; 0.03)$ and DINO obtains $44.74$ in $(0.5,\; 1.0)$. As for MoCo, the F1 drops significantly under both $(0.01,\; 0.03)$ and $(0.5,\; 1.0)$.

\setlength{\tabcolsep}{0.015cm}{\begin{table*}[h]
\vskip -0.45in
		\begin{center}
			\footnotesize
			\begin{tabular}{c| c | c | c c c c c c c c c c c c c c c c}
				\toprule
				\multirow{2}{*}{Model} & \multirow{2}{*}{Pretrain Data}  & \multirow{2}{*}{Attack Data} & \multicolumn{4}{c}{SupervisedMI} & \multicolumn{4}{c}{Variance-onlyMI} & \multicolumn{4}{c}{EncoderMI}  & \multicolumn{4}{c}{PartCrop} \cr
		\cmidrule(lr){4-7} \cmidrule(lr){8-11} \cmidrule(lr){12-15} \cmidrule(lr){16-19} 
                &  &  & Acc & Pre  &  Rec & F1 & Acc & Pre  &  Rec & F1 & Acc & Pre  &  Rec & F1  & Acc & Pre  &  Rec  & F1 \\ 
				\midrule
				\midrule
				\multirow{6}{*}{\textit{MAE}} &
				\multirow{2}{*}{\textit{CIFAR100}} 
				& \textit{CIFAR10} &  50.29 & 50.65 & 22.46 & 31.12  & 49.78 & 49.75 & 44.38 & 46.91  & 50.05 & 55.81 & 0.48 & 0.95 & \textbf{56.54} & \textbf{57.48} &  \textbf{49.32} & \textbf{53.08} \\
				& & \textit{Tinyimagenet} &  50.39 & 72.41 & 1.26 & 2.48  & 50.05 & 50.02 & \textbf{93.34} & \textbf{65.13} & 50.21 & 5.30 &  52.06 & 9.62   & \textbf{55.63}  & \textbf{81.22} & 13.84 & 23.65 \\
				\cmidrule(lr){2-19}
				& \multirow{2}{*}{\textit{CIFAR10}} 
				& \textit{CIFAR100} & 50.00 & 50.00 & 0.22 & 0.43  & 50.19 & 50.77 & 12.56 & 20.14 & 49.92 &	49.95 &	\textbf{98.06} & \textbf{66.18} & \textbf{54.19} & \textbf{52.69} & 80.94  &  63.82 \\
				&& \textit{Tinyimagenet} &  50.25 & 50.71 & 17.82 & 26.37  & 49.98 & 49.99 & \textbf{66.00} & \textbf{56.89} & 49.91 & 39.53 & 0.34 & 0.67 & \textbf{52.13} & \textbf{99.51} & 4.14 &  7.95\\
				\cmidrule(lr){2-19}
				& \multirow{2}{*}{\textit{Tinyimagenet}}
				& \textit{CIFAR100} & 49.88 & 47.78 & 2.58  & 4.90  & 49.58 & 49.65 & 60.02 & 54.34 & 49.99 & 0.04 & 40.00 & 0.08 & \textbf{55.51} &	\textbf{54.71} & 	\textbf{64.06} & \textbf{59.02}  \\
				&& \textit{CIFAR10} & 49.99 & 46.15 & 0.12  &  0.24  &49.53 & 49.58 & 55.66 & 52.44& 49.62 & 2.58 & 43.58 & 4.87 & \textbf{54.46} &	\textbf{53.85}	& \textbf{62.42} & \textbf{57.82} \\
				\midrule 
				\multirow{6}{*}{\textit{DINO}} &
				\multirow{2}{*}{\textit{CIFAR100}} 
				& \textit{CIFAR10} & 49.49 & 46.44 & 6.66 & 11.65  & 51.73 & 55.16 & 18.50 & 27.71 &\textbf{ 57.54} & 54.81 & \textbf{79.78} & \textbf{64.98} & 55.22 & \textbf{56.97} & 	39.88 & 46.92\\
				& & \textit{Tinyimagenet} &49.86 & 49.90 & 71.20 & 58.67  & 50.20 & 50.44 & 23.20 & 31.78 & 50.00 & 50.00 & \textbf{100.00}  & \textbf{66.66} & \textbf{55.82} & \textbf{63.46} & 26.96 & 37.84 \\
				\cmidrule(lr){2-19}
				& \multirow{2}{*}{\textit{CIFAR10}} 
				& \textit{CIFAR100} & 50.02 & 50.10 & 9.72 & 16.28  & 50.27 & 50.18 & 75.08 & 60.16 & 49.99 & 50.00 & \textbf{99.90} & \textbf{66.64} & \textbf{54.86} & \textbf{69.69} & 15.82 & 25.78 \\
				&& \textit{Tinyimagenet} & 50.00 & 50.00 & \textbf{99.74} & \textbf{66.61}  & 53.87 & 56.82 & 32.24 & 41.14 & 50.11 & 50.09 & 98.80 & 66.48 &  \textbf{57.23} & \textbf{69.59}	& 25.68 & 37.52  \\
				\cmidrule(lr){2-19}
				& \multirow{2}{*}{\textit{Tinyimagenet}}
				& \textit{CIFAR100} & 50.02 & 50.01 & 98.54 & 66.34  &50.00 & 50.00 & 98.28 & 66.28 & 50.00 & 	50.00	& \textbf{100.00} & \textbf{66.66} & \textbf{52.38} & \textbf{51.72} &	80.60 & 63.00 \\
				&& \textit{CIFAR10} & 49.74 & 49.85 & 89.66 & 64.07  & 50.80 & 50.42 & 96.98 & 66.34 & 50.00 & 	50.00	& \textbf{100.00} & \textbf{66.66} & \textbf{54.00}  & \textbf{52.31}  & 	90.72  & 66.35 \\
				\midrule 
				\multirow{6}{*}{\textit{MoCo}} &
				\multirow{2}{*}{\textit{CIFAR100}} 
				& \textit{CIFAR10} & 49.98 & 49.97 & 30.56 & 37.93  & 49.30 & 49.27 & 47.06 & 48.14 & 54.38 & 52.43 & \textbf{94.48} & 67.44  &  \textbf{78.61}  & 	\textbf{78.31}	 &  79.14 & \textbf{78.72} \\
				& & \textit{Tinyimagenet} &  47.77 & 44.66 & 18.64 & 26.30  & 49.84 & 49.92 & 97.70 & 66.08 & 50.00  & 50.00  & \textbf{100.00} & 66.66  & \textbf{52.88} & \textbf{50.75} &	99.88  & \textbf{67.30} \\
				\cmidrule(lr){2-19}
				& \multirow{2}{*}{\textit{CIFAR10}} 
				& \textit{CIFAR100} & 50.04 & 50.81 & 2.50 & 4.77  & 49.87 & 49.86 & 46.60 & 48.17 & 55.88 & 63.54 & 27.58 & 38.46 & \textbf{74.32} &	\textbf{87.78} &	\textbf{56.50}  & \textbf{68.75} \\
				&& \textit{Tinyimagenet} & 49.84 & 49.88 & 64.58 & 56.29  & 49.58 & 49.77 & 93.06 & 64.85 & 51.44 & \textbf{92.35}	& 3.14 & 6.07 & \textbf{53.69} & 51.92 & \textbf{99.74} & \textbf{68.29} \\
				\cmidrule(lr){2-19}
				& \multirow{2}{*}{\textit{Tinyimagenet}}
				& \textit{CIFAR100} & 49.69 & 49.79 & 76.32 & 60.26  & 49.76 & 49.87 & \textbf{93.84} & \textbf{65.13} & 50.77 & 64.98 & 3.34 &  6.35 &\textbf{56.48} & \textbf{100.00} & 12.96 &  22.95 \\
				&& \textit{CIFAR10} & 49.93 & 49.74 & 13.74 & 21.53  & 49.31 & 49.63 & \textbf{93.08} & \textbf{64.74} & 52.09 & 64.30 & 9.40 & 16.40 & \textbf{62.94} & \textbf{99.84} & 25.92 & 41.16 \\
				\bottomrule
			\end{tabular}
		\end{center} 
  \vskip 0.05in
    \caption{Compare PartCrop with three baselines in shadow setting. Acc: Accuracy, Pre: Precision, Rec: Recall} 
  \label{tab:shadow-res}
\end{table*}}

\subsection{Results in the Shadow Setting}~\label{sec:shadow setting}
We also conduct experiments in a shadow setting where an adversary has no knowledge about training data and test data, but attacks it via training an attacker on a public dataset. Specifically, for simplicity, we use the default data split, \eg, 50\% known data of CIFAR100, to train an attacker, and then attack a victim encoder trained by TinyimageNet (or CIFAR10), inferring their 50\% unknown data.

As shown in Tab~\ref{tab:shadow-res}, we perform cross-dataset membership inference for all the models and compare the results with SupervisedMI, Variance-onlyMI, and EncoderMI. We see that SupervisedMI produces around 50\% for accuracy, failing in all situations. For Variance-onlyMI, it fails to infer effectively in most cases where approximately 50\% accuracy is achieved.  For EncoderMI, similar to \textit{partial setting}, it fails to work on MAE. On DINO, only EncoderMI trained on CIFAR100 successfully attacks CIFAR10 while the rest all fail. On MoCo, there are still some failure cases but the whole performance is improved compared to DINO.
In contrast, though PartCrop achieves inferior performance compared with EncoderMI under certain conditions, PartCrop succeeds in all cross-dataset attacks and produce most of the best results in various cross-dataset settings. This strongly demonstrates the generalization of PartCrop to datasets with distribution discrepancy compared to EncoderMI.

\subsection{Potential Tapping}
In this subsection, we aim to further explore PartCrop's potential. First, we consider the impacts of stronger data augmentations on PartCrop as these are potential improvements for self-supervised models. Then we verify the generalization of PartCrop to more self-supervised models and datasets. Next we evaluate PartCrop across more self-supervised paradigms to demonstrate its broad effectiveness.

\subsubsection{Impacts of Data Augmentation} 
\begin{figure}[t]
\centering{\includegraphics[width=0.95\linewidth]{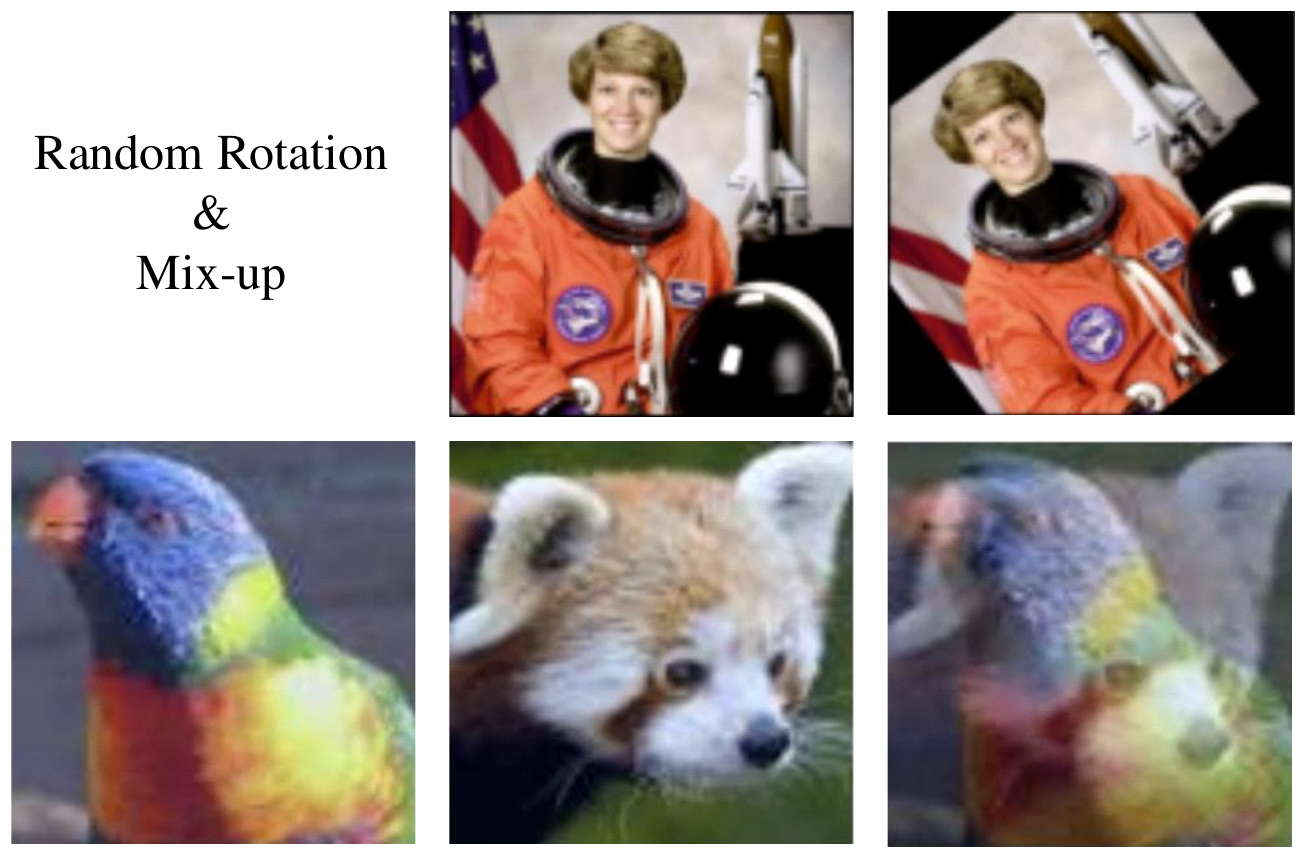}}
	\caption{Illustration of random rotation and mix-up. The first row is random rotation. The second row is mix-up~\cite{zhang2018mixup}.}
	\label{fig:Illustration}
\end{figure}

Data augmentation plays a vital role in self-supervised learning. Although experiments have demonstrated the favorable performance of PartCrop under various data augmentation strategies employed in the three self-supervised models, we introduce additional stronger data augmentations to further validate the superiority of PartCrop. Specifically, we explore three types of data augmentations including random image rotation [-180$^{\circ}$, 180$^{\circ}$] (intra-image operation), mix-up~\cite{zhang2018mixup}~\footnote{Mix-up obtains a combination of two images with a coefficient. It is used in contrastive learning~\cite{kim2020mixco}. Code is \href{https://github.com/Lee-Gihun/MixCo-Mixup-Contrast}{here}.} (inter-image operation), and adversarial learning~\cite{szegedy2014intriguing, biggio2013evasion}~\footnote{Adversarial learning is also used in contrastive learning~\cite{kim2020adversarial}. Code is \href{https://github.com/Kim-Minseon/RoCL}{here}.} (data training manner). In~\cite{kim2020mixco} and~\cite{kim2020adversarial}, we find that both methods are built upon MoCo. Hence, we adopt MoCo as the baseline and use their official code to train on CIFAR100. Regarding random image rotation ([-180$^{\circ}$, 180$^{\circ}$]), we incorporate it into the original data augmentation recipe of MoCo. The results are presented in Tab~\ref{tab:data aug}. The introduction of extra data augmentation has minimal impact on SupervisedMI as it essentially relies on random guessing. In contrast, these data augmentations generally lead to a certain degree of decrease in attack accuracy for EncoderMI, Variance-onlyMI~\footnote{We notice a slight improvement in Variance-onlyMI when adding random rotation, which is because Variance-onlyMI follows Label-onlyMI~\cite{choquette2021label} to adopts random rotation, and unintendedly mimics target model to augment images to generate membership feature, which could potentially resemble that of augmented training images.}, and PartCrop. This outcome is reasonable as these data augmentations mitigate overfitting to the training set. Fig~\ref{fig:Illustration} shows that random rotation can result in the loss of some image information, which is replaced with black in the four corners. Mix-up combines two images and introduces blurring, thereby impairing the perception of each image' content by self-supervised models and hindering the part-aware capability. Thus, the accuracy of PartCrop notably decreases. Adversarial learning adopts an adversarial manner to prevent overfitting in self-supervised models. Compared to PartCrop, EncoderMI is more susceptible to the effect of adversarial learning. However, PartCrop consistently outperforms EncoderMI across all data augmentations,  demonstrating the superiority of PartCrop in comparison to other methods. 

\setlength{\tabcolsep}{0.06cm}{\begin{table*}[h]
\vskip -0.45in
		\begin{center}
   \footnotesize
			\begin{tabular}{c| c c c c c c c c c c c c c c c c}
				\toprule
    \multirow{2}{*}{Attacker} & \multicolumn{4}{c}{MoCo} & \multicolumn{4}{c}{Random Rotataion}  & \multicolumn{4}{c}{Mix-up} & \multicolumn{4}{c}{Adversarial learning} \cr
		\cmidrule(lr){2-5} \cmidrule(lr){6-9} \cmidrule(lr){10-13} \cmidrule(lr){14-17}
                   & Acc & Pre  &  Rec  & F1 & Acc & Pre  &  Rec & F1 &  Acc & Pre  &  Rec & F1 & Acc & Pre  &  Rec & F1 \\
				\midrule
				\midrule
				 \textit{SupervisedMI} & 49.94	& 49.42	& 5.16 & 9.34 & 49.98 & 49.98 & 60.66 &  54.80 & 50.18 & 50.74 &  12.30 & 19.80 & 50.36 & 50.49 & 6.14 & 10.94  \\
                    \textit{Variance-onlyMI} & 50.94  & 50.85 & 56.06  & 53.33  & 51.01 & 51.47 & 35.22 & 41.82  & 50.20 & 50.20 & 50.12 & 50.16 & 50.07  & 50.12 & 30.56 & 37.97   \\
				 \textit{EncoderMI} & 57.29 &	58.14 &	59.12 & 58.62	 & 54.92 &	56.21	& 44.52 &  49.68 & 52.82 &\textbf{56.96} &	23.08 & 32.8 & 51.80 & 	51.88 &	49.60 & 50.71  \\
				 \textit{PartCrop} & \textbf{77.20} & \textbf{82.93} &	\textbf{67.84} & 	\textbf{74.63}  &   \textbf{73.85} &	\textbf{81.66} &	\textbf{61.52}  & \textbf{70.17} & \textbf{56.94} &	55.35 & \textbf{71.76} & \textbf{62.50}& \textbf{68.93} &	\textbf{65.98} & \textbf{78.16} & \textbf{71.56} 	\\
				\bottomrule
			\end{tabular}
		\end{center}
  \vskip 0.05in
   \caption{Compare PartCrop with three baselines on extra data augmentation. Acc: Accuracy, Pre: Precision, Rec: Recall} 
   \label{tab:data aug}
\end{table*}}

\subsubsection{Model and Dataset Generalization} As PartCrop is superior over EncoderMI, especially on masked image modeling methods, to further validate the effectiveness, we conduct experiments on another two models, \ie, CAE~\cite{chen2022context} (MIM) and iBOT~\cite{zhou2021ibot} (combine CL and MIM) using CIFAR100. As shown in Tab~\ref{tab:generalization}, In CAE, PartCrop (58.27\% Acc) outperforms EncoderMI (49.97\% Acc) by 8\%. In iBOT, PartCrop (60.03\% Acc) outperforms EncoderMI (51.73\% Acc) by 8\%. These results further verifies the generalization of PartCrop on models over EncoderMI.

On the other hand, the evaluation datasets used currently are mostly single-object images. To verify the effectiveness of PartCrop on multiple-object image datasets, \eg, MS COCO~\cite{lin2014microsoft}, we conduct attack on Long-seq-MAE~\cite{hu2022exploring} pretrained on COCO. In Tab~\ref{tab:generalization}, PartCrop obtains 55.15\% Acc while EncoderMI (50.13\%) completely fails. This is reasonable as EncoderMI is designed for CL, \eg, MoCo, which often trains on single-object images. It is worth noting that this type of experiment is absent in EncoderMI and by contrasting, we show the generalization of PartCrop on datasets as well as potentially more significant value in reality application.

\setlength{\tabcolsep}{0.06cm}{\begin{table}[h]	
			\footnotesize
			\begin{tabular}{c| c | c c c c}
				\toprule
				\multirow{2}{*}{Method} & \multirow{2}{*}{Attacker} & \multicolumn{4}{c}{CIFAR100}\cr
		\cmidrule(lr){3-6} 
                   & & Acc & Pre  &  Rec & F1 \\
				\midrule
				\midrule
				\multirow{4}{*}{CAE} 
				 & SupervisedMI & 50.24  & 50.25	& 47.60 & 48.89 \\
                    & Variance-onlyMI & 50.04  & 	50.03 & 62.40  & 55.53 \\
				 & EncoderMI &  49.97 & 49.98 & \textbf{85.18} & \textbf{63.00}  \\
				 & PartCrop &  \textbf{58.27}  & \textbf{59.53}	& 51.66 & 55.32 \\
				\midrule 
				\multirow{4}{*}{iBOT}
				& SupervisedMI &  49.94	& 42.50	&  0.34 & 0.67  \\
    & Variance-onlyMI & 49.44 & 49.57	& \textbf{64.30} & 55.98  \\
				& EncoderMI & 51.73 & 52.25	 &	40.04  & 45.34  \\
				& PartCrop & \textbf{60.03} & \textbf{60.17} & 	59.34  & \textbf{59.75} \\
				\midrule
				\multirow{4}{*}{\makecell[c]{Long$-$seq MAE \\ (MS COCO)}}
				& SupervisedMI  &  49.17 &  49.26 &	55.82 & 52.34  \\
    & Variance-onlyMI & 50.26 & 50.41 & 31.92 & 39.09 \\
				& EncoderMI & 50.13  & 50.65	& 10.16 & 16.92  \\
				& PartCrop & \textbf{55.15} & \textbf{54.99} &	\textbf{56.80}  & \textbf{55.88} \\
				\bottomrule
			\end{tabular}
   \caption{Results on extra self-supervised models.}
  \label{tab:generalization}
		\vspace{-8pt}  
\end{table}}

\setlength{\tabcolsep}{0.14cm}{\begin{table}[h]		
			\footnotesize
			\begin{tabular}{c| c | c c c c c c c c c c c c}
				\toprule
				\multirow{2}{*}{Method} & \multirow{2}{*}{Attacker} & \multicolumn{4}{c}{CIFAR100}\cr
		\cmidrule(lr){3-6} 
                   & & Acc & Pre  &  Rec & F1 \\
				\midrule
				\midrule
				\multirow{4}{*}{SWaV} 
				 & SupervisedMI & 49.97  & 49.88	& 12.78 & 20.35  \\
     & Variance-onlyMI & 50.22  & 50.19 	& 59.12 & 54.29 \\
				 & EncoderMI & 53.50  & 53.79 & 49.68 & 51.65  \\
				 & PartCrop &  \textbf{68.98}  & \textbf{69.06}	& \textbf{68.76} & \textbf{68.91} \\
				\midrule 
				\multirow{4}{*}{VICReg}
				& SupervisedMI & 49.38 & 49.42 & 53.00 & 51.14 \\
    & Variance-onlyMI &  49.79 & 49.83	& \textbf{59.88} & \textbf{54.39}  \\
				& EncoderMI & 50.56 & 50.95 & 50.55 & 50.75  \\
				& PartCrop & \textbf{55.31} & \textbf{80.55} & 14.00 & 23.85 \\
				\midrule
				\multirow{4}{*}{SAIM}
				& SupervisedMI & 50.30	&  50.24 & 61.58 & 55.33 \\
    & Variance-onlyMI & 50.35 & 50.42 & 42.38 & 46.05 \\
				& EncoderMI & 50.72 & 50.55 & 65.66 & 57.12 \\
				& PartCrop & \textbf{56.74} & \textbf{55.60} & \textbf{66.86}  & \textbf{60.71} \\
				\bottomrule
			\end{tabular}
   \caption{Results on other self-supervised paradigms.}
  \label{tab:unified}
		\vspace{-8pt} 
\end{table}}

\subsubsection{Evaluation on Other Paradigms} \label{unified}
Besides contrastive learning and masked image modeling, we additionally consider three representative (or recently proposed) models trained by extra self-supervised paradigms including clustering method (SwAV~\cite{caron2020unsupervised}), information maximization (VICReg~\cite{bardes2021vicreg}), and auto-regressive methods (SAIM~\cite{qi2023exploring}), and conduct experiments on CIFAR100. As reported in Tab~\ref{tab:unified}, PartCrop obtains 68.98\% Acc for SwAV, 55.31\% Acc for VICReg, and 56.74\% Acc for SAIM, outperforming EncoderMI (53.50\%, 50.56\%, and 50.72\%). For the attack success, we conjecture that these models may possess characteristics resembling part-aware capability. Studies in~\cite{chen2022intra} empirically proves that the representation space tends to preserve more equivariance and locality (part), supporting our speculation. On the other hand, these results indicate that PartCrop is a more unified method than EncoderMI in the field of visual self-supervised learning.

Additionally, we illustrate PartCrop's efficacy with less adversary knowledge, \ie, the proportion of the known part of a target dataset, show PartCrop's complementarity to other methods, demonstrate its effectiveness on image datasets from other domains (\eg, medical), and evaluate its inference runtime. The experiments are put in App~\ref{app:Potential Tapping}.

\section{Defenses}
Although our primary focus is to propose a new attack method, there are some commonly used defense methods for MI, \ie, early stop and differential privacy. For the integrity of our work, we follow the existing research~\cite{liu2021encodermi, jia2022badencoder, liu2022poisonedencoder, liu2022stolenencoder} and conduct an experimental verification on whether these methods are effective against our newly proposed method. We perform inference attack and classification task~\footnote{We use pretrained model to extract feature and feed it into a learnable linear classification layer following~\cite{liu2021encodermi}.} on CIFAR100. In our experiments, we find that they are not very effective, thus we newly introduce a defense method called shrinking crop scale range that can better defend against PartCorp. We organize all results in Tab~\ref{tab:all} for better comparison.

For \textbf{Early Stop (ES)}, we train MAE and MoCo with 800 epochs compared to vanilla (1600 epochs) and train DINO for 400 compared to vanilla (800 epochs)~\footnote{We do not train DINO 1600 epochs in this section as it makes DINO overfitting on training set, which leads to inferior task performance on test set. Specifically, 800 epochs produce 57.1\% classification accuracy while 1600 epochs produce 47.0\%.}. Tab~\ref{tab:all} shows that ES is generally effective to decrease the 
inference accuracy for all models but reduces the classification accuracy by a large margin for all models. For \textbf{Differential Privacy (DP)}~\cite{dp_abadi2016, shokri2015privacy}, we use Opacus and follow the official hyperparameters~\footnote{Please see \url{https://github.com/pytorch/opacus/blob/main/tutorials/building_image_classifier.ipynb}} to implement our DP training. Tab~\ref{tab:all} shows that DP enhances defense capability of self-supervised models, especially for MoCo.
On the other hand, DP also incurs significant utility loss, especially for DINO. Hence, more effective methods are expected and thereby we propose SCSR.

\setlength{\tabcolsep}{0.18cm}{\begin{table}[h]
   \footnotesize
			\begin{tabular}{c| c | c c c c | c }
				\toprule
				\multirow{2}{*}{Model} & \multirow{2}{*}{Defense} & \multicolumn{4}{c}{Attack} & Task \cr
		\cmidrule(lr){3-6} \cmidrule(lr){7-7} 
                   & & Acc ($\downarrow$) & Pre  &  Rec & F1 & Acc($\uparrow$) \\
				\midrule
				\midrule
				\multirow{4}{*}{MAE} 
				 & Vanilla & 58.4 &   57.7  & 60.0 & 58.8 & 48.3 \\
				 & ES & 56.9 & 60.3  & \underline{44.3} & \underline{51.1}  & \underline{45.1} \\
				 & DP & \underline{54.5} &   \textbf{53.4}  & 80.0 & 64.0 & 13.2 \\
        & SCSR &  \textbf{54.3} & \underline{59.6}  & \textbf{39.2} & \textbf{47.3} & \textbf{46.8} \\	
    \midrule 
				\multirow{4}{*}{DINO} 
				 & Vanilla & 58.1 & 59.4  &  48.3 & 53.3  & 57.1 \\
				 & ES & 57.0 & 58.5 & \textbf{51.1} & \textbf{54.5} & \underline{49.8} \\
				 & DP &  \textbf{54.4} & \textbf{53.3}  &  \underline{71.4} & \underline{61.0}  & 11.7 \\
        & SCSR &  \underline{55.1} & \underline{53.6} &  74.5  & 62.3 & \textbf{50.9} \\
				\midrule
				\multirow{4}{*}{MoCo} 
				 & Vanilla & 77.2 &   82.9   & 67.8  & 74.6 & 44.0 \\
				 & ES & 74.6 & 82.9   & \textbf{62.7} &  71.4 & \textbf{41.4} \\
				 & DP & \textbf{50.0} &   \textbf{50.0}   & 100.0  & \textbf{63.9} & 23.1 \\
        & SCSR &  \underline{71.3} & \underline{71.1}   & \underline{68.2}  &  \underline{69.6} & \underline{35.4} \\
				\bottomrule
			\end{tabular}
   \vskip 0.05in
	 \caption{Results on defense via three methods including early stop, differential privacy, and shrinking crop scale range in self-supervised learning. Bold (\textbf{54.3}) is the best result and underline \underline{55.1} indicates the second-best result.} 
 \label{tab:all}
\end{table}}

For \textbf{Shrinking Crop Scale Range (SCSR)},
random crop is a crucial data augmentation strategy in self-supervised learning. Generally, self-supervised models are trained with a wide range of crop scale, \eg, $(0.2,\; 1)$. Our insight is that this potentially contributes to self-supervised models' part-aware capability, which is used in PartCrop. Thus, a simple idea is to shrink crop scale range by increasing the lower bound. For example, for simplicity, we can use $(0.5,\; 1)$~\footnote{In App~\ref{sec:dis}, we discuss how to select the best lower bound of SCSR with a two-stage strategy.} instead of vanilla setting $(0.2,\; 1)$. In this way, self-supervised models fail to see small patches. In Tab~\ref{tab:all} the attack accuracy decreases for all the models. When comparing the SCSR with ES and DP, SCSR produces most of the best and second-best results in terms of the privacy and utility, exhibiting greater effectiveness. However, truthfully, the absolute effect is only somewhat satisfactory, particularly for MoCo. Hence, further research is required to explore more effective defense methods against PartCrop in the future.

\section{Exploring the Impact of Scaling on PartCrop}

Currently, similar to previous work, PartCrop only performs prototype testing on small-scale image datasets and toy visual encoders. While the results are promising, there remains a gap in assessing its performance in real-world scenarios, where self-supervised encoders are considerably large and typically trained on large-scale datasets. In this section, we address this gap and extend our investigation beyond membership inference on large models. We systematically explore the impact of scaling on membership inference for self-supervised visual encoders, considering both data and model dimensions—a topic that has received limited attention.

\subsection{Experiment Setting}

\subsubsection{Large-scale Dataset} 
\textbf{ImageNet1K}~\cite{imagenet15russakovsky} is a large-scale image dataset comprising over \textit{one million} images with 1000 categories.  Compared to ImageNet1K, the image datasets currently employed in membership inference, \eg, CIFAR10 and CIFAR100~\cite{cifar_10_krizhevsky2009learning}, are small (50000 images). 
Also, ImageNet1K is widely used in real-world scenario for training self-supervised vision models~\cite{he2022masked, he2020momentum, caron2021emerging}. Hence, it is reasonable for us to select it as our experimental dataset.

\subsubsection{Implementation Details}
We use \textit{official} code and settings to guarantee that all encoder variants are well-trained, avoiding the impact of insufficient training. To conserve computational resources, we are inclined to use officially published model weights (trained for the same number of epochs) if they are available. Otherwise, we conduct our own model training, ensuring the same training epochs. For attacker, we use the default structure of PartCrop in Tab~\ref{tab:attack structure} if not specified. When training an attacker, we follow the same setting in Sec~\ref{sec:Implementation Details}.

\subsection{Data Scaling}

Self-supervised learning is famous for leveraging extensive unlabeled data. Thus, in Sec~\ref{Scale Data Size}  we explore how scaling the size of training data for a large visual self-supervised encoder affects the membership inference performance. On the other hand, considering that an adversary may illegally purchase large quantities of data from the black market, thereby scaling the size of training data for attackers is worth consideration, which we show in Sec~\ref{Scale Data Size for Attackers}.

\textbf{Victim Encoders.} Data size would vary with different scales. Since we only scale data, to prevent overfitting or underfitting, we use relatively large ResNet50 and ViT-B as victim self-supervised encoders for all experiments below.

\textbf{Remark.} For evaluation, scaling data makes establishing a fixed attack evaluation benchmark unreasonable as datasets vary. Hence, we have to construct different evaluation benchmarks tailored to various data scaling settings. Though benchmarks are different, meaningful comparisons remain feasible. This is primarily because the benchmarks constructed for evaluation all stem from ImageNet1K~\footnote{Previous work~\cite{liu2024decade, kuhn2013applied, caruana2006empirical} demonstrates that performance comparison between subdatasets sampled from the same dataset is feasible, contingent upon maintaining the representativeness and statistical properties of the sampled subdatasets.}. Moreover, to ensure robustness and reliability in the evaluation process, we conduct three repeated experiments for each scaling setting. In this way, any variability introduced by random seeds or stochastic processes could be mitigated.

\subsubsection{Scale Data Size for Visual Encoders} \label{Scale Data Size} 

In this part, we aim to answer the following question primarily:\\
\begin{center}
\small
\fcolorbox{black}{gray!7}{\parbox{.95\linewidth}
{\textbf{Question 1: How secure are the self-supervised encoders trained by scaled datasets against membership inference?}}}
\end{center}
\;\\

Intuitively, a seemingly reasonable answer might be that the inference performance of PartCrop \textit{continuously} decreases as encoders are required to memorize more data, alleviating the overfitting. However, our experiments imply that this answer is not fully right. Below we first describe how we conduct this type of scaling experiments and then present the results.

\textit{Stage 1: Encoder training with scaled dataset}s. To train a self-supervised encoder with datasets of different scales, we construct two smaller datasets including \textbf{ImageNet1K-10\%} and \textbf{ImageNet1K-1\%} based on ImageNet1K. ImageNet1K-10\% is constructed by randomly sampling 10\% of images from each category of ImageNet1K (by maintaining the same number of categories to exclude the influence of discrepancy on category). ImageNet1K-1\% follows the same way. Then we train self-supervised encoders on ImageNet1K-10\% and ImageNet1K-1\% with the same training setting as encoders trained on ImageNet1K.

\begin{figure}[t]
\centering{\includegraphics[width=0.9\linewidth]
{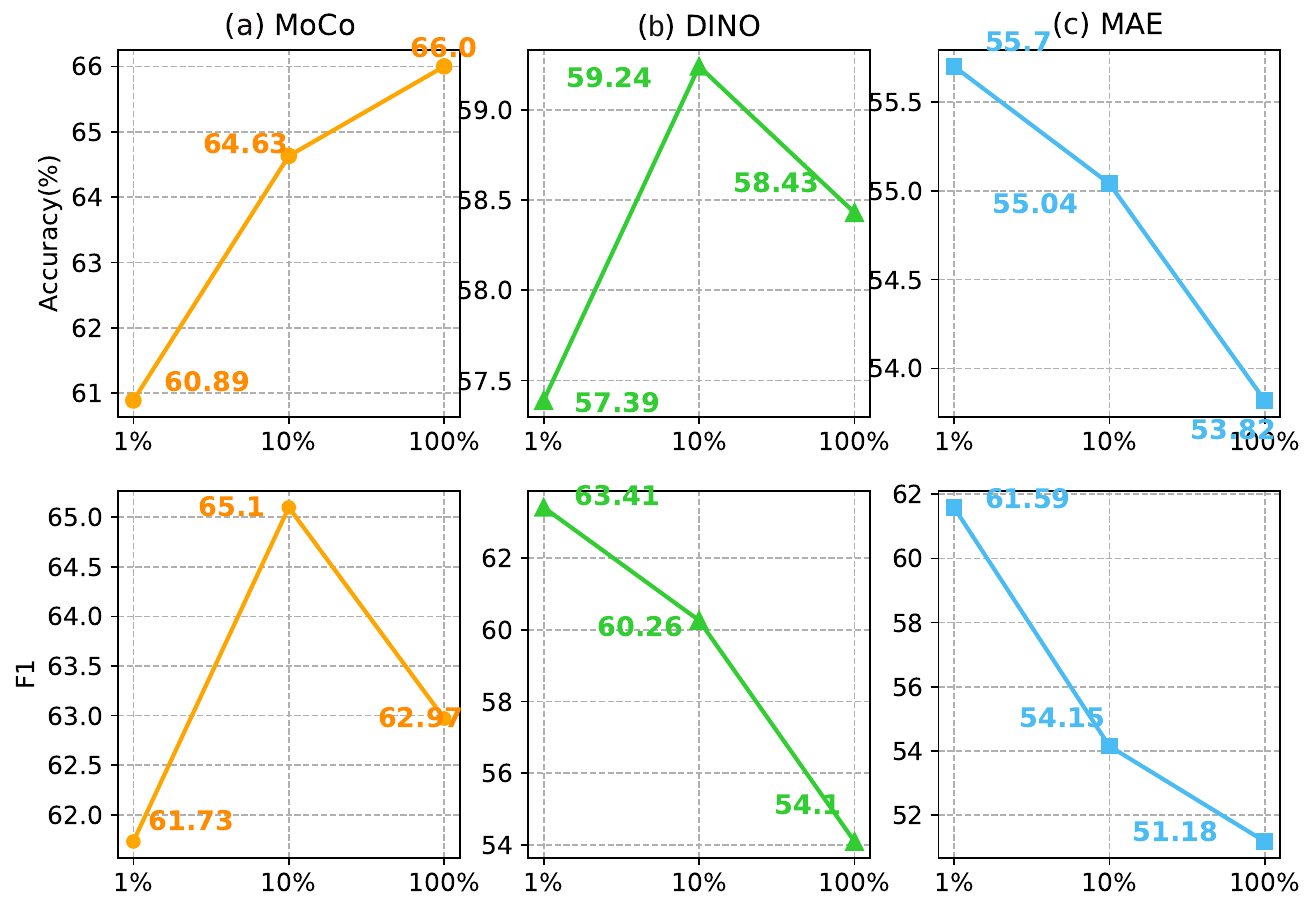}}
\caption{The results of scaling training data for self-supervised encoders.} 
\label{fig:scale encoder data}
\end{figure}

\textit{Stage 2: Attacker training and evaluation.} Next, to examine the impact of the size of the encoder's training data on the performance of membership inference, we create attack evaluation benchmarks from the corresponding ImageNet1K-1\%, ImageNet1K-10\% and ImageNet1K datasets, respectively. For instance, given the constraint that an adversary may not have access to the entire sensitive dataset used for training, following previous work~\cite{zhu2022safety}, we randomly sample 5000 images from training set and val set of ImageNet1K-1\%, respectively, to constitute the training set for attackers. Then, we randomly sample 5000 images from the left training set and val set of ImageNet1K-1\% as val set for attack evaluation. ImageNet1K-10\% and ImageNet1K follow the same process. We repeat Stage 1 and Stage 2 three times and report the averaged results. 

As presented in Fig~\ref{fig:scale encoder data}, in MoCo, enlarging training data continuously enhances the inference performance of PartCrop. We also observe that there is an improvement when scaling training data from 1\% to 10\% for DINO, but the inference performance eventually drops as we further scale the training data. The decreasing trend is more salient in both Acc and F1 for MAE. This may be due to different self-supervised paradigms. Recall that MoCo is an instance-level contrastive learning method and using more data requires MoCo to enhance its instance-level discriminative capability~\cite{he2020momentum}. Previous research~\cite{zhu2023understanding, chen2022intra} demonstrates that the enhanced discriminability could endow MoCo with stronger part-aware capability, thereby magnifying the perceived disparity between the training and test images and enabling PartCrop to produce better inference performance. This characteristic is also established for DINO in Fig~\ref{fig:scale encoder data} (b) ($1\% \rightarrow 10\%$). Nevertheless, as the dataset size increases, MoCo's generalization continues to improve, while its discriminability remains stagnant due to fixed model size. In that time, the performance of MoCo may eventually drop like DINO. To prove it, we perform PartCrop on MoCo trained on ImageNet22K (combining ImageNet21K~\cite{ridnik2021imagenet21k} and ImageNet1K~\cite{imagenet15russakovsky}) that has around 13M images. PartCrop produces 51.40\% accuracy, significantly inferior to 66.0\% (ImageNet1K), verifying our speculation. \\

\begin{figure}[t]
\centering{\includegraphics[width=0.9\linewidth]{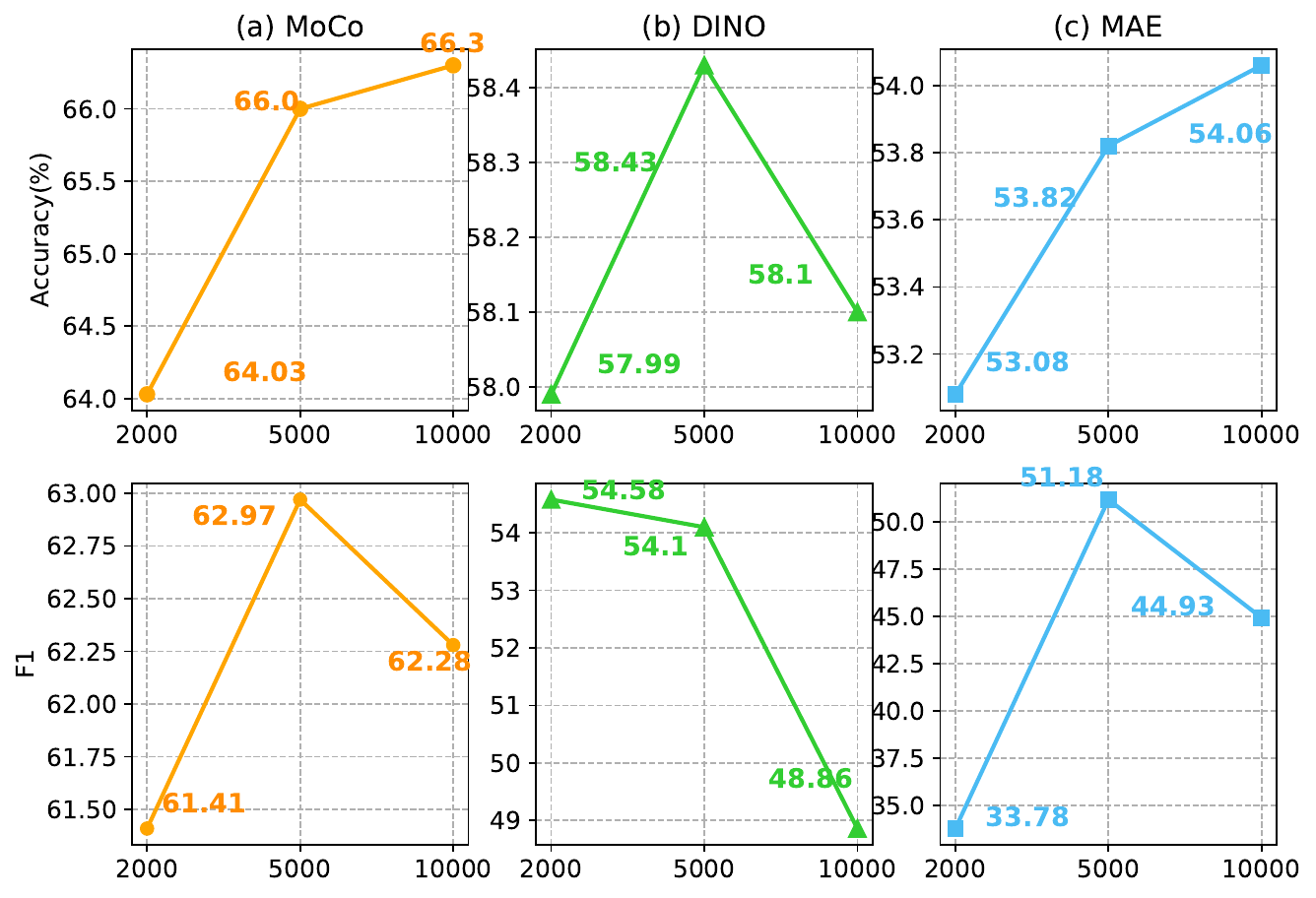}}
\caption{The results of scaling training data for attackers on self-supervised encoders.} 
\label{fig:scale attack data}
\end{figure}

\begin{center}
\small
\fcolorbox{black}{gray!7}{\parbox{.95\linewidth}{\textbf{Answer 1: Scaling the size of training data for self-supervised encoders reduces the risk of membership inference. However, this reduction occurs more gradually in the contrastive learning paradigm.}}} 
\end{center}

\subsubsection{Scale Data Size for Attackers} \label{Scale Data Size for Attackers}

In this part, we aim to answer following question primarily:\\

\begin{center}
\small
\fcolorbox{black}{gray!7}{\parbox{.95\linewidth}{\textbf{Question 2: How scaling the size of an attacker's training images affect membership inference?}}} 
\end{center}
\;\\

To answer Question 2, similarly, we sample 2000 (5000, or 10000) images from training set and val set of ImageNet1K, respectively, to constitute a training set for training attackers. Then, we randomly sample 5000 (5000, or 5000)~\footnote{We only scale data size for training set while keeping the size of val set for fair evaluation.} images from the left training set and val set of ImageNet1K, respectively, as a val set for attack evaluation. The whole process is repeated three times and we report the averaged results. 

As shown in Fig~\ref{fig:scale attack data}, regarding MoCo and MAE, PartCrop continuously improves their inference performance with the size of training data scaled. The improvement shows the benefits of scaling attackers' training data. In the case of DINO, the inference performance improves when enlarging training data at the beginning and is still higher than that of 2000 when the size is 10000. Though we fail to observe continuous benefit of scaling on DINO eventually, PartCrop indeed enjoys the benefit of scaling at the beginning.
Therefore, to some extent, scaling the size of training data for attackers is worthwhile. Besides, the improved performance on both MoCo and MAE indicates that scaling training data is a widely helpful method to enhance inference performance. \\

\begin{center}
\small
\fcolorbox{black}{gray!7}{\parbox{.95\linewidth}{\textbf{Answer 2: Scaling the size of training images for attackers is an effective method to enhance attack performance.}}} 
\end{center}
\;\\

\begin{figure}[t]
\centering{\includegraphics[width=0.9\linewidth]{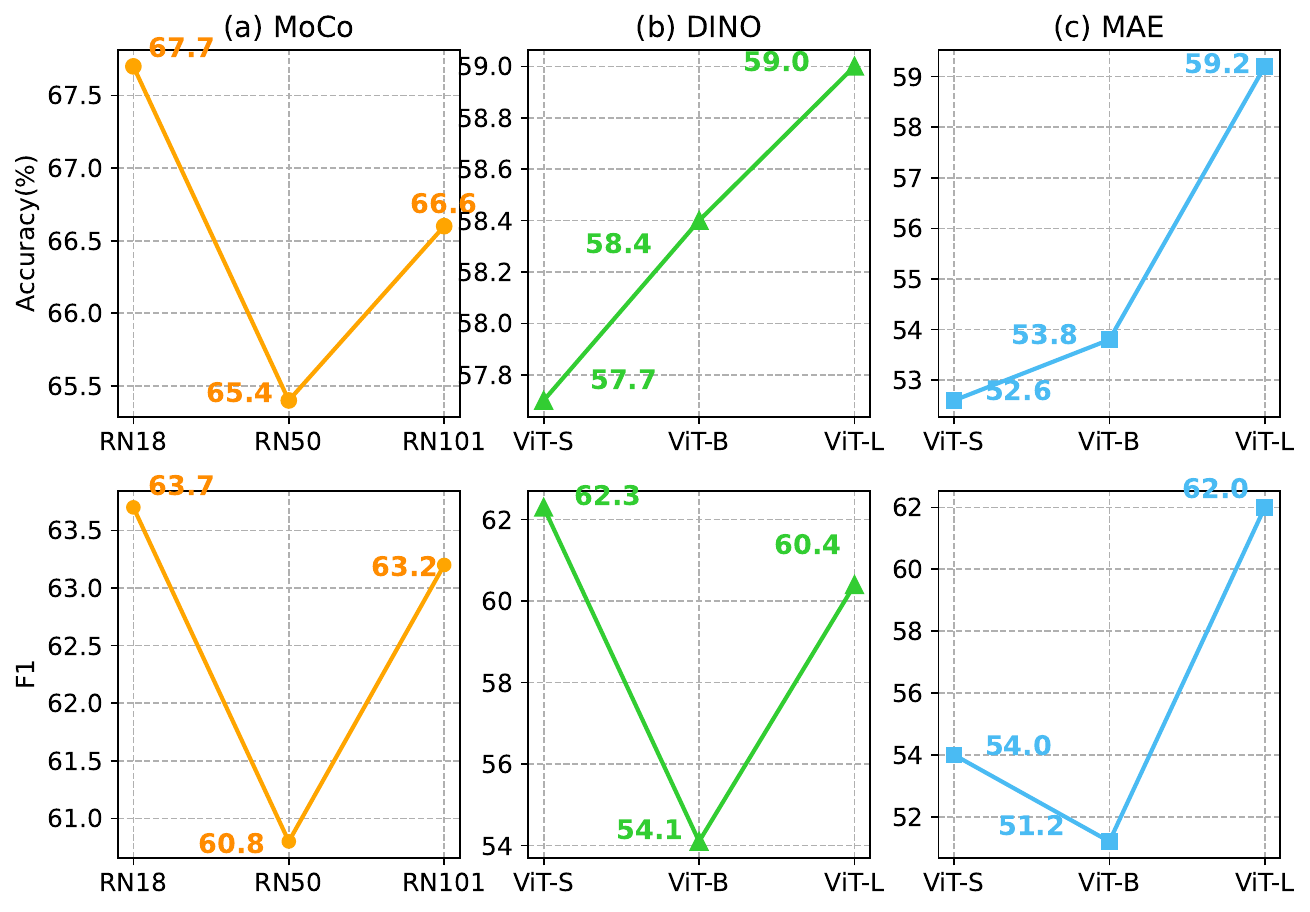}}
	\caption{The results of scaling the size of self-supervised encoders for PartCrop, respectively. RN18: ResNet18} 
	\label{fig:scale-encoder}
\end{figure}

\subsection{Model Scaling} \label{sec:scale model size}

We primarily perform model scaling from two aspects, \ie, \textit{self-supervised encoders} and \textit{employed attackers} as they are both highly related to the inference performance. Note that in this subsection, encoders are all pretrained on ImagNet1K. For a fair comparison, we establish a fixed attack evaluation benchmark since we only scale encoders and attackers. Below we describe how we construct such an evaluation benchmark.

\textbf{Evaluation Benchmark.} We randomly sample 5000 images from training set and val set of ImageNet1K, respectively, to constitute the training set for training attackers. Then, we sample 5000 images from the left training set and val set of ImageNet1K, respectively, as val set for attack evaluation. To avoid the influence of random seeds, we repeat the whole process three times and report the averaged result. These evaluation benchmarks are adopted for all experiments below.

\subsubsection{Scale Victim Self-supervised Encoders} 

In this part, we aim to answer the following question primarily:\\ 

\begin{center}
\small
\fcolorbox{black}{gray!7}{\parbox{.95\linewidth}{\textbf{Question 3: How secure are the scaled self-supervised encoders against membership inference?}}} 
\end{center}
\;\\

For each self-supervised method, we use three encoder variants with gradually increasing model size to show how the inference performance varies. For these variants, we follow the default training recipe of each variant and keep the same training epochs to avoid its effect. The results are presented in Fig~\ref{fig:scale-encoder}. We see that PartCrop generally increases the attack accuracy with the size of self-supervised encoder scaled. This trend is quite intuitive and reasonable as increasing 
model size endows self-supervised encoders more capability to memorize a greater amount of information from the training images.\\

\begin{center}
\small
\fcolorbox{black}{gray!7}{\parbox{.95\linewidth}{\textbf{Answer 3: Scaling self-supervised encoders generally elevate the risk of membership inference. Thus, when we enjoy better task performance introduced by scaling model size, it necessitates us to consider the additional privacy risk brought by such scaling.}}}
\end{center}
\;\\

\subsubsection{Scale Attackers}
Scaling an attacker is also a potential strategy to improve the performance of membership inference. Instead of directly increasing the number of parameters of an attacker, we aim to 
decouple the effects of two distinct scaling strategies \ie, being wider (increase channel number) and being deeper (add more layers). By doing so, our study may provide more sophisticated and useful prior for adversaries who can flexibly define her own attacker structure.




\textbf{Remark.} It is important to highlight that, as far as we know, there hasn't been extensive research on the optimal design of attackers to enhance inference performance. To some extent, our findings may serve as valuable experimental references for future researchers in this area.

In the following, we aim to answer the question primarily:\\

\begin{center}
\small
\fcolorbox{black}{gray!7}{\parbox{.95\linewidth}{\textbf{Question 4: What changes in inference performance of PartCrop occur when the attacker widens or deepens? How this type of scaling study inspires adversaries?}}} 
\end{center}
\;\\

To answer \textbf{Question 2}, besides the default attacker in Tab~\ref{tab:attack structure}, we further consider four different attacker variants including \textbf{\textit{narrow}}, \textbf{\textit{wide}}, \textbf{\textit{shallow}}, and \textbf{\textit{deep}}, to give a comprehensive understanding of the impact of scaling attackers. Our modification criteria is to bring as few modifications as possible to default attacker, via which we aim to reduce the impact of other factors brought by such modifications. Hence, Tab~\ref{tab:variants} shows the details about how the four variants are derived from the default attacker. Additionally, considering that our primary focus is on scaling attacker, we use a widely-used ResNet50 and ViT-B as our victim encoders, employ the same training setting as the default attacker, and report the averaged results.

\setlength{\tabcolsep}{0.04cm}\begin{table}[h]
	\centering
 \footnotesize
\begin{tabular}{c | l}
		\toprule
		Variants & Modifications \\
		\midrule
		\textbf{\textit{Narrow}}   & replacing $d=512$ with $d=1024$  \\
  \textbf{\textit{Wide}}   & replacing $d=512$ with $d=256$ \\
 \textbf{\textit{Shallow}}   & merge Layer 2 and 3 into ``Linear(d, d/4)+ReLU" \\
  \textbf{\textit{Deep}}  & add ``Linear(d/2, d/2)+ReLU" between Layer 2 and 3  \\
		\bottomrule
	\end{tabular}
  \caption{The details of the four attack variants.}
   \label{tab:variants}
\end{table} 

\begin{figure*}[t]
\centering{\includegraphics[width=1\linewidth]{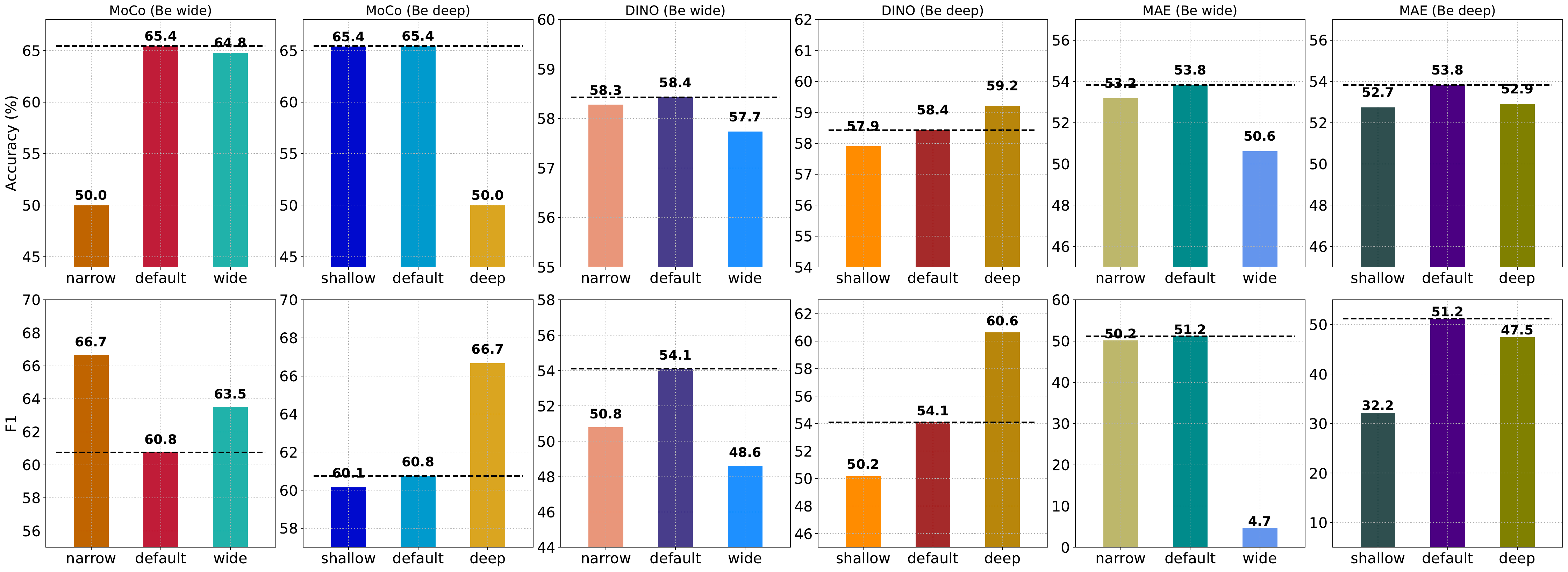}}
 \vskip -0.05in
	\caption{The results of different attacker variants using PartCrop on MoCo, DINO, and MAE. We use large ResNet50 for MoCo and ViT-B for DINO and MAE. We report the averaged result for each variant.}
	\label{fig:partcrop-scale-attacker}
\end{figure*}

Fig~\ref{fig:partcrop-scale-attacker} illustrates the results where inference performance of different attacker variants varies significantly, highlighting the importance of exploring the design of attacker structures. In MoCo, it seems that \textit{default} generally performs the best among all attacker variants. We also see that DINO prefers a deeper attacker as it increasingly improves the inference performance (both accuracy and F1). When it comes to MAE, almost all variants achieve similar performance. Interestingly, we find that for all encoders, the medium attacker, \textit{default}, basically performs the best or second best. Thus, opting to use a moderately sized attacker is comparatively more suitable.

\begin{figure}[h]
\hspace{0.16cm} \footnotesize \textit{Default} 
\centering{\includegraphics[width=1\linewidth]{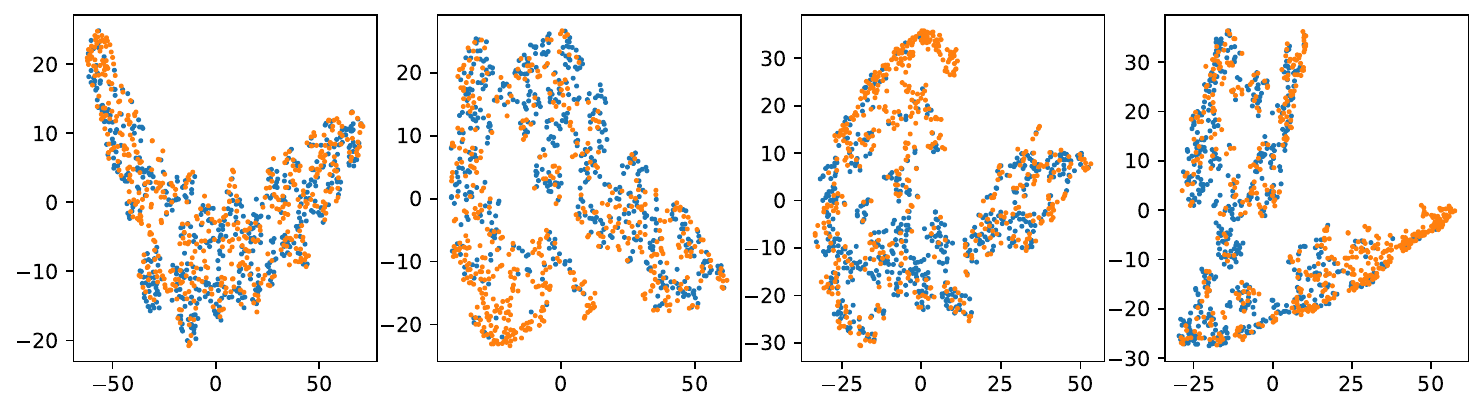}}
\hspace{0.32cm} \footnotesize \textit{Narrow} 
\centering{\includegraphics[width=1\linewidth]{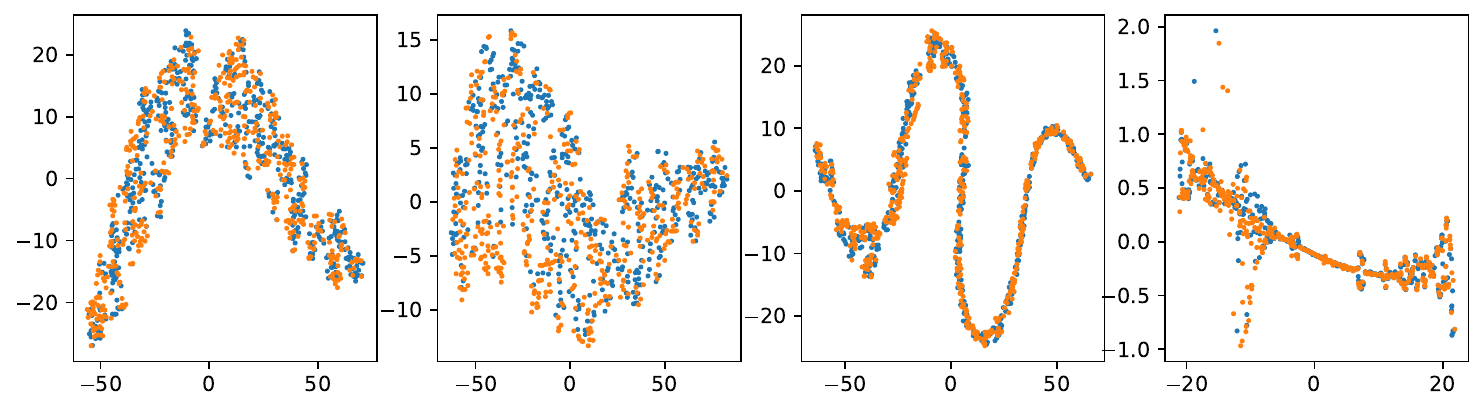}}
\hspace{0.24cm} \footnotesize \textit{Deep}
\centering{\includegraphics[width=1\linewidth]{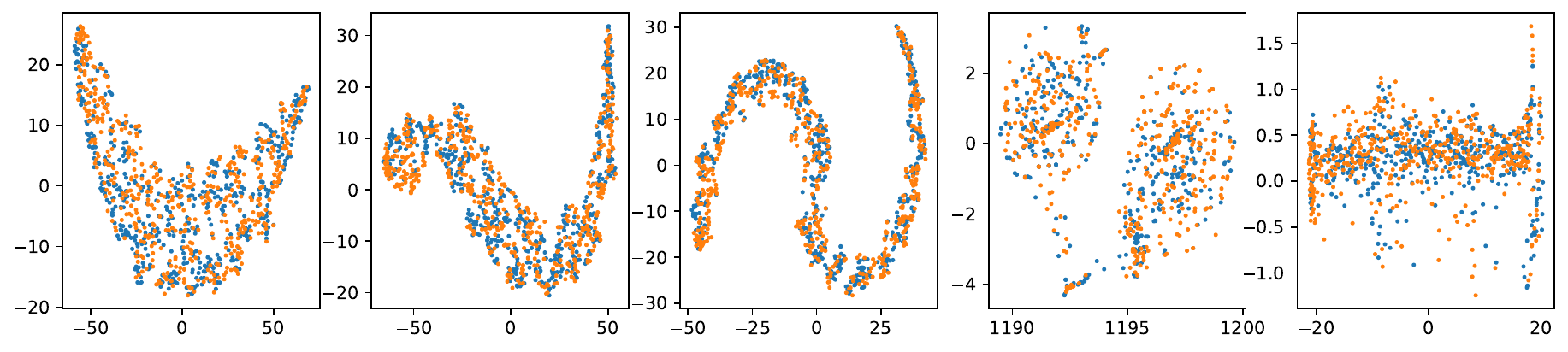}}
\caption{The t-SNE visualization of the input and sequential layer activations. The first column is the visualization of input and the rest are that of activation.} 
\label{fig:t-NSE}
\end{figure}

 
In Fig~\ref{fig:partcrop-scale-attacker}, we also notice that there is a significant degeneration (reducing to random guessing) for \textit{narrow} and \textit{deep} in MoCo, which prompts us to delve into the underlying reasons. As shown in Fig~\ref{fig:t-NSE}, we visualize the learned feature of each attacker variant. Specifically, we sample 500 member images and 500 non-member images from val set of our evaluation benchmark, adopt t-SNE~\cite{van2008visualizing} to visualize the inputs and high-dimensional representations from all activation layers (\ie, ReLU) of \textit{default}, \textit{narrow}, and \textit{deep} variants, respectively.

We see that the visualization of all inputs (in the first column) is similar, with a `V' shape. This is reasonable as all inputs come from the same sample and are built following the same process~\footnote{There are also a slight difference in shape, which is negligible and may be caused by random seeds when loading and cropping patches.}. This result excludes potential issues of inconsistent inputs. However, after that, \textit{narrow} varies dramatically where all representations are clustered around a curve and lose distinguishability. We think the current dimension $d$, especially in the third and fourth columns, is too small for \textit{narrow} to learn sufficient information, and the ReLU function further reduces the representation's discernibility due to its suppression of negative value. For \textit{deep},  the value of ordinate and abscissa in the third, fourth, and fifth columns varies dramatically. For example, the value of abscissa abruptly shifts from $25$ to $1200$, which may imply that the feature learned in \textit{deep} is not stable. This may be due to lacking normalization in PartCrop attacker. 

To alleviate these issues, we accordingly introduce two structural improvements to PartCrop, rendering it as a scalable attack method. Specifically, we replace ReLU with Tanh to preserve information from negative value, and introduce Root-Mean-Square normalization (RMSNorm~\cite{zhang2019root}) to improve feature stability, which is a simple variant of L2 normalization and easier to train. We term this resulted attacker as PartCrop-v2. In Tab~\ref{tab:PartCrop-v2}, we see that only using Tanh can remarkably alleviate the issue of random guessing for \textit{narrow} as well as \textit{deep}. This may be due to the implicit normalization of Tanh which simply scales value to $[-1, 1]$. However, this manner may lead to suboptimal results and also fails to address our newly considered attack variant \textit{wide} in MAE, which also exhibits a potential tendency towards random guessing. Hence, it is necessary to introduce RMSNorm, which is more robust, and further improves attack performance. We also explore other activation functions and normalization techniques, \eg, leaky-ReLU and L2 normalization, and show the results in Tab~\ref{tab:combination}, where their performance is inferior to PartCrop-v2.


\setlength{\tabcolsep}{0.08cm}\begin{table}[h]
	\centering
 \footnotesize
\begin{tabular}{c | ccc | ccc | ccc}
    \toprule
       Attacker & \multicolumn{3}{c|}{\textit{Narrow} (MoCo)} & \multicolumn{3}{c}{\textit{Deep} (MoCo)} & \multicolumn{3}{|c}{\textit{Wide} (MAE)} \\
    \midrule
     Tanh   & \XSolidBrush & \Checkmark & \Checkmark & \XSolidBrush & \Checkmark  & \Checkmark & \XSolidBrush & \Checkmark & \Checkmark \\
     RMSNorm & \XSolidBrush & \XSolidBrush & \Checkmark  &  \XSolidBrush & \XSolidBrush & \Checkmark &  \XSolidBrush & \XSolidBrush & \Checkmark  \\
     \midrule
     Acc (\%)     & 50.0  & 65.6  & \textbf{66.2}  & 50.0 & 65.4 & \textbf{66.2} & 50.6 & 50.0 & \textbf{52.7} \\
    \bottomrule
\end{tabular}
\vskip 0.05in
 \caption{The performance of PartCrop-v2 and its ablation study.}
  \label{tab:PartCrop-v2}
\end{table} 

\textbf{Activation and Normalization Alternatives for PartCrop-v2.} Besides Tanh and RMSNorm, we also consider other activation functions and normalization choices. For activation function, we employ a improved version of ReLU called Leaky ReLU~\cite{maas2013rectifier} where the negative section allows a small gradient instead of being completely zero and a recently widely-used activation called SiLU~\cite{elfwing2018sigmoid}. For normalization, we consider two widely-used methods including L2 normalization and Layernorm~\cite{lei2016layer}. We compare the inference performance of their combinations with PartCrop-v2 on MoCo using \textit{narrow} in Tab~\ref{tab:combination}. It is shown that PartCrop-v2 produces the best performance for accuracy. \\

\setlength{\tabcolsep}{0.4cm}\begin{table}[t]
	\centering
 \footnotesize
\begin{tabular}{c | ccc } 
    \toprule
       ACC (\%) & Tanh & Leaky ReLU & SiLU \\
    \midrule
     RMSNorm   & \textbf{66.2}   & 66.0  &  66.1 \\
     L2 Norm & 50.0 & 50.0 &  50.0 \\
     LayerNorm    & 65.2  & 65.7  & 50.0 \\
    \bottomrule
\end{tabular}
\vskip 0.05in
 \caption{Inference performance of different combinations of activation function and normalization.}
  \label{tab:combination}
\end{table} 

\begin{center}
\small
\fcolorbox{black}{gray!7}{\parbox{.95\linewidth}{\textbf{Answer 4: The variation of inference performance is irregular. Generally, a moderately sized attacker is an appropriate choice as a small attacker may not sufficiently learn discernible representation while a complex one may lead to instability. Hence, we recommend our stably scalable PartCrop-v2 and suggest adversaries to consider Tanh and RMSNorm as well when designing attack structure in the future.}}}
\end{center}

\section{Conclusion}
In this article, we propose PartCrop, a unified method, to perform membership inference on a self-supervised model while having no knowledge about the algorithms and hyperparameters used. This setting is more likely to meet the reality that adversaries usually can not access to the training recipes. Specifically, by measuring the response between image and cropped parts in representation space, PartCrop produces disriminative membership feature for inference. Moreover, to defend against PartCrop, we consider two common defense methods, \ie, early stop and differential privacy, and also propose a tailored defense method called shrinking crop scale range. Extensive experiments are conducted using three computer vision datasets to verify the effectiveness of the attack and defense methods. Finally, we quantitatively study the impacts of scaling from both model and data aspects in a realistic scenario and propose a stronger and scalable PartCrop-v2 by introducing two structural improvements to PartCrop to enhance the performance of membership inference.









\section*{Declarations}

\begin{itemize}
\item Competing interests: The authors have no competing interests to declare that are relevant to the content of this article.
\item Data availability: The datasets used in this article are publicly available online.
\item Code availability: Our code is available at \url{https://github.com/JiePKU/PartCrop}.
\end{itemize}


\bibliography{sn-bibliography}

\clearpage
\begin{appendices}


\input{s-appendix}

\end{appendices}

\end{document}

%% file: s-appendix.tex
\section{Related Work} \label{app:related work}
\textbf{Membership Inference and Defense.} Currently, membership inference methods mainly include binary-classifier-based methods~\cite{Shaow_Learning, ResAdv} and metric-based methods~\cite{dp_yeom2018privacy, pre_conf_ML_Leaks, song2021systematic_entropy}. Besides classification, membership inference has been widely extended to other fields, \eg, sentence embedding~\cite{song2020information}, generative models~\cite{hayes2017logan, goodfellow2020generative}, graph model~\cite{duddu2020quantifying}, multi-modal models~\cite{hu2022m}, 
and self-supervised models~\cite{liu2021encodermi, zhu2024unified}. It is also vital to study how to defend against MI. Researchers recently consider leveraging model compression technicals to defend against MI, \eg, knowledge distillation~\cite{KD_hinton2015distilling, shejwalkar2021membership, tang2022mitigating} and pruning~\cite{Pruning_IJCAI}. Essentially, they aim to avoid target model being overfitting to training set. But Yuan \& Zhang~\cite{pruning_defeat_yuan2022membership} find that pruning makes the divergence of prediction confidence and sensitivity increase.

\textbf{Self-supervised Learning.} As a representative masked image modeling method and contrastive learning, MAE~\cite{he2022masked} directly reconstructs the masked RGB color and further boosts the downstream performance. For contrastive learning, MoCo~\cite{he2020momentum}, as one of represetatives, proposes a momentum strategy. Further, in the transformer era, DINO~\cite{caron2021emerging} explores new properties derived from self-supervised ViT. There are also works trying to understand self-supervised methods~\cite{xie2022revealing, kong2022understanding, zhu2023understanding, saunshi2022understanding, chen2022intra, zhong2022self, wei2022contrastive}. Additionally, due to the potential of SSL in leveraging massive unlabelled data, studying how to attack and protect self-supervised models is also critical~\cite{jia2022badencoder, liu2022poisonedencoder, liu2022stolenencoder, saha2022backdoor, dziedzic2022dataset, li2022demystifying, cong2022sslguard, liu2021encodermi}. 

\textbf{Scaling Dataset \& Model.} Scaling model size and dataset is critical to  improve performance. Schuhmann~\etal present a dataset of $5.85$ billion CLIP-filtered image-text pairs~\cite{schuhmann2022laion}, 14x larger than LAION-400M~\cite{schuhmann2021laion}. Tan~\etal emphasize that larger models tend to achieve superior performance~\cite{tan2019efficientnet}. CLIP~\cite{radford2021learning} is pretrained on a massive image-text dataset, allowing the model to learn a broad understanding of the relationships between different concepts in images and text. 
Kirillov~\etal propose Segment Anything Model (SAM )~\cite{kirillov2023segment}.

\section{Self-supervised Model} \label{app:Self-supervised Model}
\textbf{MAE~\cite{he2022masked}} is one of representative \textit{masked image modeling} methods that predicts masked parts of an image from the visible parts. MAE involves an encoder $\mathcal{E}$ and a decoder $\mathcal{D}$. They are both constructed by Vision Transformer~\cite{DosovitskiyB0WZ21}. Given an image, MAE first partitions it into patches, $ \rm{P} = \{\mathsf{P}_1, \mathsf{P}_2, \dots,
\mathsf{P}_N\}$. They are then randomly masked with a given mask ratio, \eg, 75\%, partitioning $\rm{P}$ into two parts, visible (unmasked) patches $\rm{P}_{v}$ and masked patches $\rm{P}_{m}$. The visible patches $\rm{P}_{v}$ are fed into encoder $\mathcal{E}$ to derive representation. The output representations along with learnable tokens $\rm{T}$ are served as input for decoder $\mathcal{D}$ to reconstruct the masked patches $\rm{P}_{m}$. The objective function is formulated as:
\begin{equation}
\small \mathcal{L} = MSE(\mathcal{D}(\mathcal{E}(\rm{P}_{v})\,, \rm{T})\,, \rm{P}_{m}) \,,
\end{equation}
where $MSE$ is mean square error loss which measures the absolute distance between the reconstruction output of decoder $\mathcal{D}$ and RGB value of masked patches $\rm{P}_{m}$ in normalized space. After training, the encoder $\mathcal{E}$ is preserved for downstream tasks \eg, classification, while the decoder $\mathcal{D}$ is abandoned.

\textbf{DINO~\cite{caron2021emerging}} is representative among \textit{contrastive learning} methods, which maximizes the agreement of two augmented views from the same image. DINO contains two encoders, student $\mathcal{E}_{s}$ and teacher $\mathcal{E}_{t}$, constructed by Vision Transformer~\cite{DosovitskiyB0WZ21}. DINO augments a given image with different strategies, generates two views $v_{1}$ and $v_{2}$, and feed them into two encoders. Specifically, $v_{1}$ is fed into the student $\mathcal{E}_{s}$ followed by a projector $\mathcal{P}_{s}^{j}$ and a predictor $\mathcal{P}_{s}^{d}$. And $v_{2}$ is fed into the teacher $\mathcal{E}_{t}$ and a projector $\mathcal{P}_{t}^{j}$. The two output representations are going to agree with each other by the objective formulated as:
\begin{equation}
\small \mathcal{L} = CE\{\mathcal{P}_{s}^{d}(\mathcal{P}_{s}^{j}(\mathcal{E}_{s}(v_{1})))\,, \mathcal{P}_{t}^{j}(\mathcal{E}_{t}(v_{2}))\} \,.
\end{equation}
$CE$ is cross-entropy loss. Minimizing the loss brings the two representations close in representation space. After training ends, the student encoder $\mathcal{E}_{s}$ or the teacher encoder $\mathcal{E}_{t}$ is preserved for downstream tasks while the rest are abandoned. We use the student encoder $\mathcal{E}_{s}$ in this research.  

\textbf{MoCo~\cite{he2020momentum}} is well-known among \textit{contrastive learning} methods in CNN era, which brings two positive samples that are from the same image with different augmentation strategies close and pushes away negative samples from different images. MoCo consists of two encoders (a vanilla encoder $\mathcal{E}$ and a momentum encoder $\mathcal{E}_{m}$) and a dynamic dictionary $\Phi$. $\mathcal{E}$ and $\mathcal{E}_{m}$ have the same structure, \eg, ResNet~\cite{resnet_he2016deep}, and output feature vectors. But $\mathcal{E}_{m}$ updates slowly due to \textbf{e}xponential \textbf{m}oving \textbf{a}verage (EMA) strategy. For clarity, we denote feature vectors from $\mathcal{E}$ as \textit{queries} $q$ and feature vectors from $\mathcal{E}_{m}$ as \textit{keys} $k$. $\Phi$ is a queue that pushes the output vectors ($k$) from $\mathcal{E}_{m}$ in current mini-batch in queue and pops representations of early mini-batch. In MoCo, an image is firstly augmented to generate two different samples $s_{1}$ and $s_{2}$. Then $s_{1}$ is fed into $\mathcal{E}$ and $s_{2}$ is fed into $\mathcal{E}_{m}$, producing two representations ($q_{s}$ and $k_{s}$) respectively. They are regarded as a positive pair as they are from the same image while $q_{s}$ and representations in $\Phi = \{\mathsf{k}_1, \mathsf{k}_2, \dots,
\mathsf{k}_N\}$ are negative pairs. Then InfoNCE loss~\cite{oord2018representation, he2020momentum, chen2020improved} is employed for training:
\begin{equation}\label{eq:moco}
\small \mathcal{L} = -\log \frac{\exp(q_{s} \cdot k_{s} / \tau)}{\exp(q_{s} \cdot k_{s}/ \tau + \sum_{i=1}^{N}(q_{s} \cdot \mathsf{k}_i)/ \tau)} \,.
\end{equation}
$\tau$ is the temperature coefficient.  In each iteration, $\Phi$ pops representations $\mathsf{k}$ of early mini-batch and pushes $k_{s}$ in queue as new $\mathsf{k}$. In this way, $\Phi$ updates dynamically.  After training, $\mathcal{E}$ is preserved for downstream tasks while the rest are discarded.

\section{Baseline} \label{app:Baseline}
\textbf{SupervisedMI.} SupervisedMI takes model output and label (one-hot form) as the input of attacker. Unfortunately, in self-supervised learning, label is absent. Hence, we repurpose this baseline by taking the model output as input to train attacker. Specifically, given an image $x$, we feed it into self-supervised model that outputs feature vector $\xi$. Then the feature vector $\xi$ is regarded as membership feature and fed into the attacker for training. Following previous works~\cite{ResAdv, zhu2022safety, liu2021ml}, we use a similar fully connected network as attacker.  

\textbf{Variance-onlyMI.} Generally, models are less sensitive to augmentations of training samples than of test samples. Label-only inference attack leverages this characteristic and take the collection of model-classified label from various augmented images as input for inference. However, in self-supervised learning, label is absent. To enable this attack, we repurpose this baseline by using various augmented images to compute the channel-wise representation variance as attack input instead of labels. Intuitively, member data usually produce smaller variance.  We denote this modified attack as Variance-onlyMI. Following~\cite{choquette2021label}, we use the same augmentation strategies and a similar network as attacker.

\textbf{EncoderMI.} EncoderMI is the most related work with ours, a strong baseline as it knows how the target model is trained in its setting. EncoderMI use the prior that contrastive learning trained encoder tends to produce similar feature vectors for augmented views from the same image. Hence, it is necessary to know training recipe, especially the data augmentation strategies, to generate similar augmented views that appear in training. To perform inference, given an image $x$, EncoderMI creates $n$ augmented views using the same data augmentation strategies as the victim self-supervised model. The $n$ views are denoted as $x_{1}, x_{2}, \dots, x_{n}$. Then they are fed into the victim model to produce corresponding feature vectors ($\eta_{1}, \eta_{2}, \dots, \eta_{n}$). To obtain membership feature of $x$, EncoderMI calculates the similarity between $n$ feature vectors by the formula:
\begin{equation}
\small M(x) = \{S(\eta_{i}\,, \eta_{j}) \; | \; i\in [1\,,n], j\in [1\,,n], j > i \} \,,
\end{equation}
where $S(\,,)$ is the similarity function. We use cosine similarity and vector-based attacker setting as it performs the best in ~\cite{liu2021encodermi}. When obtaining the membership feature, EncoderMI ranks the $n\cdot(n-1)/2$ similarity scores in descending order and feeds them into an attacker, a fully connected neural network. 

\textbf{The discrepancy between EncoderMI and PartCrop.} After introducing EncoderMI, it is essential to delve into the discrepancy between EncoderMI and PartCrop to elucidate our work's novelty:

$\bullet$ Firstly, in terms of granularity, EncoderMI operates at the image-level, whereas PartCrop operates at a finer-grained, part-level. Specifically, EncoderMI computes similarities between images (feature vectors), while PartCrop calculates similarity distributions between each generated object part (feature vector) and the entire image (feature map).

$\bullet$ Secondly, in terms of method, EncoderMI and PartCrop rely on different priors. EncoderMI use the prior of contrastive learning, while PartCrop use the part-aware capability in self-supervised models.

$\bullet$ Thirdly, in terms of attack scope, unlike EncoderMI, which exclusively focuses on contrastive learning, PartCrop serves as a unified approach that can be employed in  contrastive learning, masked image modeling, as well as other paradigms.

$\bullet$  Lastly, and perhaps most importantly, in terms of reality, we adopt a more realistic setting where adversaries are unaware of the training recipe (methods and training details), while EncoderMI assumes adversaries possess knowledge of the method (contrastive learning) and the associated details.

\textbf{Needed information assumed by each baseline.} The SupervisedMI and Label-onlyMI assume the victim model is supervised and necessitate model-classified labels to produce membership features. However, in the realm of self-supervised learning, labels are unavailable and the encoder only outputs image representations. EncoderMI is assumed to know training details and augmentation hyperparameters, which is unrealistic as self-supervised models often function as black-box services~\cite{Shaow_Learning} in practice. Hence, these attacks on self-supervised models lack plausibility in reality. We also discuss the differences between PartCrop and general loss/entropy/confidence-based MI methods and the challenges of attacking self-supervised models compared to supervised models in App~\ref{sec:dis}.

\section{Potential Tapping} \label{app:Potential Tapping}

\begin{figure}[t]	
\centering{\includegraphics[width=1\linewidth]{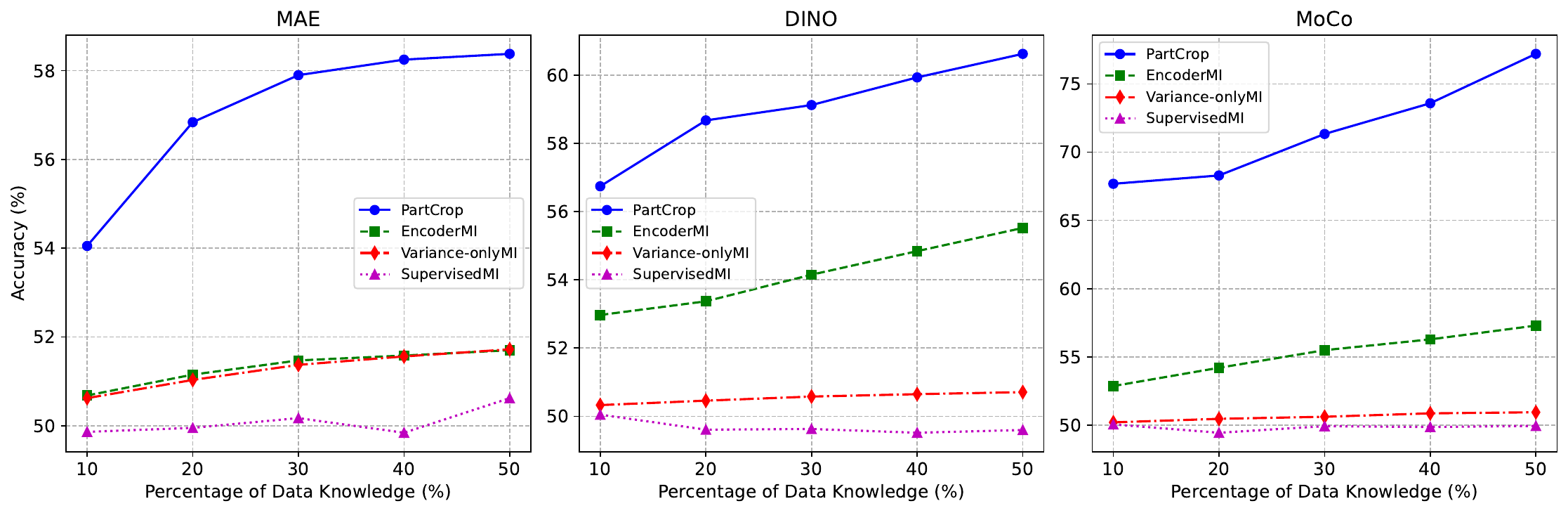}}
 \vskip 0.05in
	\caption{The illustration of a sweep experiment on the adversary's knowledge assumption from 10\% to 50\%.}
	\label{fig:ratio}
 \vskip -0.1in
\end{figure}

\textbf{Varying Adversary's Knowledge.} 
To demonstrate that PartCrop can work with limited knowledge about the samples, we perform an experiment using CIFAR100, varying the adversary's knowledge assumption from 10\% to 50\% on MAE, DINO, and MoCo, respectively, and compare the results with three baselines. As shown in Fig~\ref{fig:ratio}, the accuracy of PartCrop, EncoderMI, and Variance-onlyMI shows a steady improvement, while that of SupervisedMI varies irregularly. This is reasonable, as SupervisedMI is designed for supervised models and only performs random guesses in the self-supervised case. We also observe that PartCrop consistently outperforms EncoderMI and Variance-onlyMI under various settings of the adversary's knowledge. Moreover, though with only a 10\% knowledge assumption, PartCrop still achieves comparable accuracy (\eg, 56.74\% Acc on DINO), which outperforms that of EncderMI with a 50\% knowledge assumption (55.52\% Acc on DINO). This result highlights PartCrop's effectiveness and practicality in real-world scenarios.

\textbf{PartCrop's complementarity to other methods.} 
To validate this, we combine the membership features of PartCrop and EncoderMI to train an attacker to attack DINO using CIFAR100. This approach yields an accuracy of 62.02, further improving upon PartCrop's performance (60.62 Acc), which indicates that PartCrop is also complementary to other membership inference methods.

\textbf{Evaluation on diverse datasets.}
We evaluate PartCrop on more diverse and dynamic datasets, such as medical images, to better illustrate the effectiveness of the proposed method in various real-world scenarios. Specifically, we evaluate PartCrop on DINO, trained using the HAM10000 medical dataset~\cite{DVN/DBW86T_2018} with default settings. PartCrop achieves 56.32 Acc, outperforming EncoderMI (52.44 Acc), the strongest baseline, by a large margin. This result verifies the effectiveness and adaptability of PartCrop on diverse datasets.

\textbf{Runtime on high-resolution images.}
In real-world scenarios, images are often of high resolution. Hence, we adopt the widely-used ImageNet1K~\cite{deng2009imagenet, alexnet_krizhevsky2012imagenet} dataset with a 224x224 resolution on a large ViT-B model to evaluate PartCrop's inference runtime. The experiment is conducted on an NVIDIA 3090 (24GB) with an Intel Core i9-12900K. PartCrop's runtime is 67.8 ms per sample, faster than EncoderMI, the strongest baseline (107.8 ms per sample).

\section{Discussion}\label{sec:dis}

\textbf{Differences from the general MI methods based on model loss/entropy/confidence.} 
Though general MI methods and PartCrop both focus on inferring a model's member data, they use different characteristics of the model. General MI attacks use statistical features derived from the model's outputs. These statistical features essentially represent an overall numerical abstraction of the model's predictions. In contrast, PartCrop utilizes a subtle part-aware capability that is implicitly developed during the training process of self-supervised learning (SSL), primarily focusing on sensitivity to and understanding of local information in an image. 

\textbf{Challenges compared to attacking supervised models.} There may be three challenges in implementing an attack on a self-supervised model compared to a supervised one. First, SSL allows the model to learn from substantial unlabeled data, avoiding severe overfitting on training data, which makes inferring member data harder. Second, SSL's proxy tasks~\cite{geiping2023cookbook} are richer and more sophisticated than merely fitting labels in supervised learning (SL). Hence, a unified method is necessary. Finally, SSL's membership features require crafted designs and are not as straightforward as in SL, where features such as entropy/confidence can be used.

\textbf{A two-stage strategy to select the lower bound of SCSR.} The lower bound is a key hyperparameter for defending against PartCrop. Hence, we can use a two-stage  coarse-to-fine strategy. In the first stage, we choose a wide initial lower bound range, \eg, 0.3, 0.4, and 0.5, and assess the defense capability for each configuration. In the second stage, we further select the two best-performing lower bounds and proceed with finer adjustments between them using smaller increments, \eg, 0.02. In this way, we can fine-tune the lower bound with greater precision, ensuring optimal defense performance.

\textbf{Limitation and improvement.} For simplicity, PartCrop randomly crops patches from image to potentially yield part-containing crops with pre-define crop scale for all datasets. However, this may lead to some queries of low quality due to two potential issues: Random cropping could involve background noises, namely, patches without part of an object (noisy queries); Pre-define crop scale may not be suitable for all datasets whose image sizes are different. Considering that part is critical for attcking a self-supervised model, we use large number of crops in PartCrop to provide sufficient information to train attacker. 
Though this strategy in PartCrop is effectively verified by our experiments, it still wastes the computational resources to some extent due to noisy queries. To alleviate these issues, we could adopt an adaptive clipping strategy. For example, given an image, before clipping it, we can leverage a detector~\cite{carion2020end, liu2023grounding, wang2023detecting} (or a segmentor~\cite{kirillov2023segment, zhu2021crf, ma2024segment}) to generate object bounding boxes (object masks) as the approximate clipping scopes. Then we can crop object parts from these areas, by which we ensure that the cropped parts contain meaningful content. This adaptive clipping strategy could significantly enhance the quality of the resulting parts. Additionally, we can further develop selection criteria for the clipping ratio for different datasets based on the results of these preprocessing methods. Specifically, we can compute the average size of all objects in the dataset. If the average size is small, we should appropriately lower the upper bound of the clipping ratio to prevent the cropped parts from being too large. Conversely, we can increase its lower bound. When combining these improvements, we can refine the clipping process and ensure that the generated parts are optimized for various datasets, allowing us to consistently produce high-quality, meaningful parts to enhance the performance and training efficiency of PartCrop.